\documentclass{article}
\usepackage[utf8]{inputenc}
\usepackage[a4paper, total={6in, 10in}]{geometry}

\usepackage{amsmath}
\usepackage{amssymb}
\usepackage{pxfonts}
\usepackage{graphicx}
\usepackage{csquotes}
\usepackage{comment}
\usepackage{soul}
\usepackage{natbib}
\usepackage[english]{babel}
\usepackage[dvipsnames]{xcolor}

\providecommand{\keywords}[1]
{
  \small
  \textbf{\textit{Keywords---}} #1
}

\title{Wave based damage detection in solid structures using artificial neural networks}
\author{Frank Wuttke$^{1,*}$, Hao Lyu$^{1}$, Amir S. Sattari$^{1}$, Zarghaam H. Rizvi$^{1}$    \\
        \small $^{1}$ Geomechanics and Geotechnics Group, Kiel University, Germany \\
        \small $^{*}$ Corresponding Author: Frank Wuttke, Email: frank.wuttke@ifg.uni-kiel.de \\
}
\date{} 

\begin{document}
\maketitle

\begin{abstract}
The identification of structural damages takes a more and more important role within the modern economy, where often the monitoring of an infrastructure is the last approach to keep it under public use. Conventional monitoring methods require specialized engineers and are mainly time consuming. This research paper considers the ability of neural networks to recognize the initial or alteration of structural properties based on the training processes. The presented work here is based on Convolutional Neural Networks (CNN) for wave field pattern recognition, or more specifically the wave field change recognition. The CNN model is used to identify the change within propagating wave fields after a crack initiation within the structure. The paper describes the implemented method and the required training procedure to get a successful crack detection accuracy, where the training data are based on the dynamic lattice model. Although the training of the model is still time consuming, the proposed new method has an enormous potential to become a new crack detection or structural health monitoring approach within the conventional monitoring methods.


\end{abstract}

\section{Introduction}

For the permanent use of existing structures in urban areas, as well as for lifelines and for safety structures such as dams, the successful monitoring of structures is of the highest priority. Usually, conventional methods of structural dynamics are used to analyse the state of structures and to find existing or propagating damages, while wave based method are less used in ordinary structural dynamics. These methods are more usual in the field of non-destructive testing (NDT) \citep{Kaewunruen2006, SamanFarhangdoust2019} under use of very high excitation frequencies and shorter wave lengths. Beside the type of analysis method - structural dynamics with long wave length or ultra-sound methods in NDT with extreme short wave length - the matter of the structural analysis is based on the active analysis of excited vibrations or wave fields until the structural damage is detected. Without any knowledge of preexisting damage zones, the analysis can take lots of time. The present development shows a new strategy from a numerical case study to analyse structures under use of artificial neural networks. The artificial neural networks (ANN) \citep{SHA20071747} are used to learn the change of structural response patterns of damaged structures in opposition to non-damaged structures. 
\newline \newline
With the booming development in large-scale data generation and computation power, deep learning algorithms, especially deep convolutional networks (known as CNNs or ConvNets), are developing by leaps and bounds. Deep learning methods have been applied to various many fields, such as computer vision and natural language processing, and often outperforms conventional methods \citep{lecun2015deep}. The multi-layer structured deep models can learn patterns from data at multiple levels of abstraction. Convolution operation performs an important role in CNN layers. ConvNets are particularly suitable to learn from array-like data, such as audio, image, and video. These data are uniformly sampled from signals in spatial and/or temporal domains. In many applications, ConvNets have been used as the backbone for pattern extraction, e.g., recognising objects from images \citep{girshick2015fast}, understanding text \citep{kim2014convolutional}, and synthesising audio \citep{oord2016wavenet}. In recent years, attempts have been made to apply ConvNets to damage detection as well \citep{abdeljaber2017real, rautela2019deep}.

Deep learning methods have long been expected to be promising for wave-based damage detection \citep{AVCI2021107077}. Compared to the hand-engineered-feature-based methods, the deep-learning-based method uses deep neural networks as a feature extractor to learn representations from wave fields\citep{Guo2020Detection}. 1D CNNs and RNNs are two popular structures to recognize patterns from 1D signals. Abdeljaber et al. trained multiple 1D CNNs to detect whether damage exists at specific locations (joints) \citep{abdeljaber2017real}. Their model uses the acceleration signal at each joint as input and requires extra workload to segment the signal into frames. Considering damage detection as a binary prediction problem, i.e., predicting whether a crack exists from input data, 1D CNN, RNN, and LSTM models all can achieve high accuracy \citep{rautela2019deep}. In one following paper, the authors developed a two-stage damage detection method \citep{RAUTELA2020114189}. The method determines whether a sample is damaged or not at the first stage and then predict the location and length of the damage with another regressor network. However, the regressor network deals only with the damage that is orthogonal to the sample's surface. Khan et al. transformed the structural vibration responses in the time domain into two-dimensional spectral frame representation and then applied a 2D CNN to distinguish between the undamaged and various damaged states \citep{KHAN2019586}. Besides using CNNs for wave-based damage detection, there are also methods using different input data or a combination with other learning schemes. For example, in \citep{Gulgec2019}, the authors proposed to use 2D CNN to predict the bounding box of a crack from raw strain field. By adding noise and changing loading patterns to augment the data, the model could achieve some robustness. Nunes and her colleagues developed a hybrid strategy to detect undefined structural changes \citep{Nunes2020}. They proposed to apply a unsupervised k-means clustering method to the CNN learned features, and thus the features that are extracted from the samples with undefined changes are expected to fall out of these clusters. Our work differs from the above mentioned works by training an end-to-end model to predict both crack shapes and locations from large amounts of simulated wave fields.

The numerical treatments for the paper are done by use of a meso-scale method as it is better suited to capture the effects as initial cracking and crack propagation depending on the material parameter and the initial- \& boundary conditions without pre-definition of damaged patches, and it is also applicable for 2D and 3D problems. The Lattice Element Method is a class of discrete models in which the structural solid is represented as a 3D assembly of one-dimensional elements \citep{Wongetal2014, Rizvietal2019, Sattarietal2017}. This idea allows one to provide robust models for propagation of discontinuities, multiple crack interactions, or cracks coalescence even under dynamic loads and wave fields. Different computational procedures for lattice element methods for representing linear elastic continuum have been developed. Beside different mechanical, hydro-mechanical and multi-physical developments, the extension and basics for a new dynamic Lattice Element Method was presented in \citep{Rizvietal2018, Rizvietal2020}. This development will be used in the given paper for the health monitoring of structures. 

To perform the damage detection as a suitable numerical software, pattern indicators and specific designed neural networks are needed. The numerical simulation is realized under use of Dynamic Lattice-Element Method, where the advantage of the discontinuum method in opposition to continuum methods related to the damage detection will be discussed in the methodology section. The implemented artificial neural networks are also described in this section. Based on the considered numerical and DNN (deep neural networks) models, a case study of a 2D plane is performed to show the developments and results of the new approach.


\section{Methodology}

\subsection{Dynamic Lattice Approach}
\label{sec:2_1}

The assembly of the heterogeneous and homogeneous material will be generated by specific meshing algorithms in LEM. The Lattice Element Models with the lattice nodes can be considered as the centers of the unit cells, which are connected by beams that can carry normal force, shear force and bending moment. Because the strain energy stored in each element can be exceeded by a given threshold, the element is either removed for cracking or assigned a lower stiffness value. The method is based on minimizing the stored energy of the system. The size of the localized fracture process zone around the static or propagating crack plays a key role in failure mechanism, which is observed in various models of linear elastic fracture mechanics and multi-scale theories or homogenization techniques. Normally this propagating crack process needs a regularization, however, an efficient way of dealing with this kind of numerical problem is by introducing the embedded strong discontinuity into lattice elements, resulting
in mesh-independent computations of failure response. The generation of the lattice elements are done by Voronoi cells and Delaunay itself (\cite{Sattarietal2017, Moukarzeletal1992}). With the performance of this procedure an easy algebraic equation is generated for the static case. To develop the dynamic LEM for simulation of a propagating wave field, a more complex extension of the LEM is needed. The following solution of the dynamic LEM is solved as a transient solution in the time domain.

\subsubsection{Equation of motion}
To solve the dynamic LEM, the static LEM needs an extension of the equation of motion. The general equation of motion without the damping term is defined by
\begin{equation}
\mathbf{M}\ddot{u}+\mathbf{K}u = F(t)
\end{equation}
where $\mathbf{M}$ and $\mathbf{K}$ are the mass and the stiffness matrices terms and $F(t)$ is the applied time-dependent force. Both matrices, the mass and stiffness matrix, have to be defined in terms of the LEM definition. 

\subsubsection{Mass Matrix generation}

The mass matrix or the consistent mass matrix (CMM) is generated either by lumping the mass at the nodes or by following the variation mass lumping (VMM) scheme. The VMM scheme is also implemented in the finite element method for dynamic simulations. 
The element mass $M^{e}$ is computed using the following equation 
\begin{equation}
M_e = \int \rho \left \lfloor N_{v}^{e} \right \rfloor^{T}N_vd\Omega 
\end{equation}
If the shape functions are identical, that is, $N_{v}^{e} = N^{e}$, the mass matrix is called the consistent mass matrix (CMM) or $M_{c}^{e}$. 
\begin{equation}
M_{c}^{e}=\int_{0}^{l}\rho A\left [ N^{e} \right ]^{T}N^{e}dx=\frac{1}{4}\rho l A \int_{-1}^{1}\begin{bmatrix}
1-\epsilon \\
1+\epsilon  
\end{bmatrix}
\begin{bmatrix}
1-\epsilon & 1+\epsilon
\end{bmatrix}
d\epsilon  
\end{equation}
Where, $\rho$ is the density assigned to the Voronoi cells and $A$ and $l$ are the area and the length of the lattice elements \\
The elemental mass matrix is symmetric, physically symmetric, and complies with the condition of conservation and positively. To obtain the global mass matrix, a congruent transformation is applied. In contrast to the stiffness matrix, translational masses never vanish. All the translational masses are retained in the local mass matrix. The global transformation is achieved through the following equation.

\begin{equation}
\bar{M}_{c}^{e}=\left [  T^{e}\right ]^{T}\left [ M_{c}^{e} \right ]\left [ T^{e} \right ]
\end{equation}

\begin{equation}
M_{c}^{e}= \frac{1}{2}m^{e}\begin{bmatrix}
1 & 0 & 0 &0 \\ 
0 & 1 & 0 & 0\\ 
 0& 0 & 1 & 0\\ 
 0& 0 & 0 & 1
\end{bmatrix}
\end{equation}

\subsubsection{Element Stiffness Matrix}

The force displacement component of a truss element is given by the spring relation
\begin{equation}
\left \{ F \right \}=\left [ K \right ]\left \{ U \right \}
\end{equation}
The vectors $\left \{ F \right \}$ and $\left \{ U \right \}$ are the member joint force and member joint displacement, respectively shown in figure 2.  The member stiffness matrix or the local stiffness matrix is [K]. For a truss element it is given by

\begin{equation}
\left [ K \right ] = \frac{EA}{L}
\begin{bmatrix}
1 & 0 & -1 & 0\\ 
 0& 0 & 0 & 0\\ 
-1 & 0 & 1 & 0\\ 
0 & 0 &0  & 0
\end{bmatrix}  
\end{equation}

After applying the congruent transformation, the member stiffness matrix in global coordinates are given as 
\begin{equation}
\left [ K^e \right ]=\left [ T^e \right ]^T\left [ K \right ]\left [ T^e \right ]
\end{equation}
\begin{equation}
    K^e = \frac{E^e A^e}{L^e}\begin{bmatrix}
l^e & lm & -l^2 &-lm \\ 
 lm& -m^2 & -lm & -m^2\\ 
 -l^2& -lm & l^2 & lm\\ 
 -lm& -m^2 & lm & m^2
\end{bmatrix}
\end{equation}
Where $l = cos\phi ^e, m = sin\phi^e$ and with $\phi ^e$ as the orientation angle as shown in Figure 2.  \\
\subsubsection{Time domain solution of Equation of Motion}
The equation of motion for the linear system of equations is solved with the Newmark beta method due to its unconditional stability. The displacement and the velocity terms for the next time step are calculated as follows:
\begin{equation}
    u_{t}=u_{t-\Delta t}+\Delta t\dot{u}_{t-\Delta t}+\left ( \frac{1}{2} - \beta \right )\Delta t^{2}\ddot{u}_{t}
\end{equation}
\begin{equation}
    \dot{u}_t = \dot{u}_{t-\Delta t}+(1-\gamma )\Delta t\dot{u}_{t-\Delta t}+\gamma \Delta t\ddot{u}_t
\end{equation}
We follow the average acceleration approach with  $\beta =\frac{1}{4}$ and $\gamma = \frac{1}{2}$ \\
The Newmark beta method solves the algebraic form of the equation of motion (EOM) of undamped forced vibration at the end time interval $t+\Delta t$
\begin{equation}
    F_{t+\Delta t}=\mathbf{M}\ddot{u}_{t+\Delta t} + \mathbf{K}u_{t+\Delta t}
\end{equation}
The stiffness and the mass matrices are computed in the following fashion to reduce in the form of equation (9)  
\begin{equation}
    \mathbf{\hat{K}} = \mathbf{K}+a_{0}\mathbf{M}
\end{equation}
Where $\hat{K} $ is the effective stiffness matrix and $a_{0}=\frac{6}{\gamma \Delta t^2}$ \\
Similarly, the effective load vector at time $t+\Delta t$ is calculated as in (17).
\begin{equation}
    \hat{F}_{t+\Delta t}= F_{t+\Delta t}+\mathbf{M}(a_0 u_t + a_2\dot{u}_t + a_3\ddot{u}_t)
\end{equation}
Here, $a_2=\frac{1}{\gamma \Delta t}$ and $a_3=\frac{1}{2\gamma }$. \\
The above simplification leads to the algebraic form.

\begin{equation}
\left \{\hat{F} _{t+\Delta t}\right \} = \left [ \mathbf{\hat{K}} \right ]\left \{ U_{t+\Delta t} \right \}
\end{equation}

From the above equation, displacement of each node is calculated for every time step.\\
The natural frequency of the system is calculated as given below.
\begin{equation}
    \omega ^2[M]\Phi =[K]\Phi
\end{equation}
\begin{equation}
    \omega ^2=eig([M]^{-1}[K])
\end{equation}

The detailed description of the theory and implementation of the dynamic Lattice-Element method with validation and verification of the method by analytical and numerical benchmarks is given in \citep{Rizvietal2018, Rizvietal2020}.

\subsection{Wave field identification by Convolutional Neural Networks}

The basic idea of deep learning system damage detection in the given case of propagating wave fields is the identification of wave field patterns respective of the change in wave field patterns during the damage evolution. To apply the idea, the excitation and receiver points keep constant during the monitoring process. 

The damage evolution process covers the initial static case on the given plate. After a change of surrounding, static stress condition damages on different positions in the plate area can be created depending on the stress condition and the material parameter. Before and after a damage scenario, a small strain wave field is excited to propagate through the plate. Because of the damage / crack, within the plate the pattern of the propagating wave field will be modified. 

The interaction of the wave field within the crack is essential for identifying the correct wave field. Under the assumption of an open crack, neglecting shear slipping and crack growth under dynamic loads, the crack will produce a mode conversion and a scattering of the propagating wave field. That phenomena is studied in \citep{Rizvietal2018, Rizvietal2020}. It becomes obvious that the transient solution provides that phenomena. 

For future application of the DNN-Damage-Detection of real structures, the virtual tool has to be optimized and needs sufficient training data. The numerical data should provide the base of the training data of the CNN.

\subsection{Training of 1D CNN-Based Detector for Damage Detection}
\subsubsection{The Damage Detection Dataset}
\paragraph{An Overview of Training Data Set - The Wave Propagation Data}
The wave field data was produced by numerical simulation on virtual 2D plates as study examples. After the application of a Dirac input, the elastic wave propagates from the source point into the plate. Adding source points at a different location of the plate generates different wave patterns, which are recorded by a set of receiver points. Particularly when the plate has a crack inside, the wave patterns are very different from the one in non-crack plates. Because of the crack, the wave field shadow as well as the reflection becomes visible and change the pattern of the wave field around the damaged region.

The plate has fixed boundary conditions at the bottom and no constrains at other sides. The wave source can change its location around the plate boundary. Receivers that record wave displacement are assumed to be along the free boundary as well as on the inner surface of the plate.
\newline \newline
The sequential measures of time-dependent displacement amplitudes are used as the data source for the planned damage detection network. The instantaneous load added at a chosen excitation point causes high displacement amplitudes at the wave field and decreases rapidly after several time steps to the wave coda at a smaller strain level. The observable surface wave front, as well as at interference of back scattered and reflected wave field is the result of the wave propagation and reflection in the plate. The described phenomenon are clearly visible in an example in Section 3.1.
\newline\newline
In this paper, the whole content of the wave field, the initial wave front, the coherent part and the diffusive part will be used for the damage detection in the time domain. There is no selection respective analysis of harmonic wave modes in the part coherent wave field \citep{Wuttke2012a} or application of the interferometric method at the code in the diffusive part \citep{Wuttke2012b} of the wave field yet.

\begin{figure}[h!]
\centering
\includegraphics[scale=0.4]{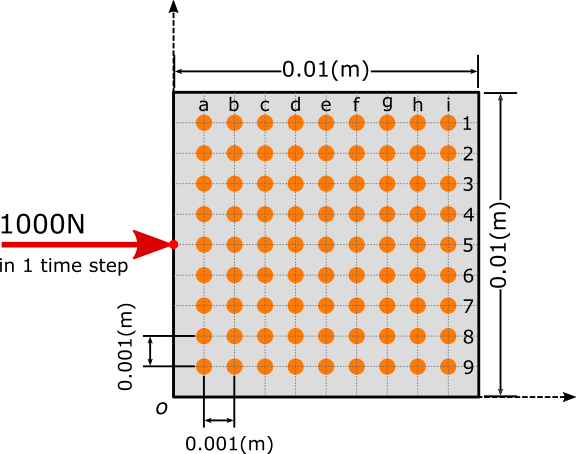}
\caption{The receiver arrangement on the surface of a plate.}
\label{fig:sample_config}
\end{figure}

\paragraph{The Damage Detection Dataset} 
The dataset is generated by running the numerical simulation repeatedly for randomly generated plates with or without a crack. In this study, the size of all plates is set to $0.01(m) \times 0.01(m)$ and the lower-left plate corner is defined as the Cartesian coordinate system origin. The wave field is generated by an initial Dirac impulse (in 1 times step span). The simulation runs for 2000 samples with time steps by $1e-9$ seconds. With the excitation of the wave field, the recording at all receiver points starts as the base of the deep detection network. During the method development, possible negative effects by larger displacement values on deep learning model are analysed. The plate, receivers and excitation points are shown in Figure \ref{fig:sample_config}. The resulted displacement wave field consists of time histories in the X- and Y-direction at 81 receiver positions with 2000 time steps.

To validate the damage detection method with randomly generated cracks, the crack itself is described by 3 parameters, i.e., crack length $l$, orientation $\alpha$, and start position $(x, y)$, with their values being randomly chosen from the following description,
\begin{align}
    l& \in (0, \frac{1}{2} \min (e_x, \, e_y)], \\
    \alpha& \in [0, 360], \\
    x &\in [s_x, \, e_x-s_x], \\
    y &\in [s_y, \, e_y-s_y],
\end{align}
where $e_x$ and $e_y$ are the length of the sample edges along the x-axis and y-axis, $s_x$ and $s_y$ are the distance between two receivers in the X-axis and Y-axis (see Figure \ref{fig:sample_config}). If one randomly generated crack stretches out of the sample plate, the excess part is discarded. The plate particles that correspond to the crack are marked as removed for the Lattice-Element model calculation.

\begin{figure}[h!]
\centering
\includegraphics[scale=0.8]{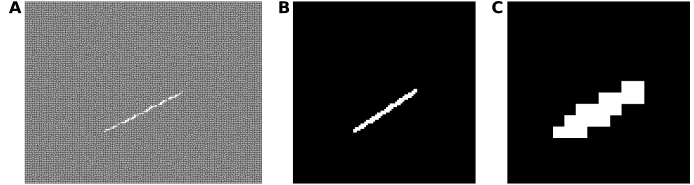}
\caption{Plate with crack and its binarised labels at different resolutions. (A). The generated sample with one crack inside in particle view. (B). The 100x100 binary label image for the plate (white pixels indicate the existence of the crack). (C). The 16x16 binary label image for the plate.}
\label{fig:bin_2res}
\end{figure}

To start the identification scenario, i.e., detecting the damage in a given plate, a binary image of the damage was provided as its label for comparison between the identified structure and original structure. The binary image covers the plate's surface and indicates the location where a crack exists. The label image can be obtained originally of an 100x100 resolution, where each pixel covers an area similar to the size of a particle. When the model is adjusted to refine or enlarge  predictions, the resolution of the label image can be changed accordingly. Figure \ref{fig:bin_2res} gives out two label images of different resolutions for the same plate. The image of 16x16 pixels is resized and binarized from the image of 100x100 pixels. We use the image of the reduced resolution (16x16) as the supervision signal for training the detector network. As the proposed model does predictions for each pixel, the label image of 16x16 resolution restricts the problem scale while still maintaining the model's applicability.

The numerical simulation randomly generates plate particles and cracks within the plate domain using pre-defined boundary conditions. These randomnesses in sample generation and crack generation are called \textit{Type-N} samples. Additionally, plates without any crack inside are also generated randomly as reference samples, marked as \textit{Type-R}. It's also possible to generate samples of different plates with similar cracks (\textit{Type-S}) or the same plates with different cracks (\textit{Type-C}). In this work, 3040 samples for training and 320 samples for testing were generated. In the training dataset, nearly half of the cases are Type-N; Type-R, Type-S, and Type-C cases consist of the remaining portion of the training data. The testing data consists of 144 Type-N samples, 8 Type-R samples, 160 Type-C samples, and 8 Type-S samples. The Type-C samples are generated from 10 different samples and 16 different cracks for each sample (see Figure \ref{fig:groundtruth-valid-origin} in the Appendix). Among all test samples, we intentionally generated 7 random samples, and each one has its counter case in the training dataset in terms of the same crack (no-crack cases are excluded in this case). It is worth emphasizing that these samples are not repeated ones. Because of the randomness in the generation process, the diversity of the interior particles and their wave field patterns is ensured.

\subsubsection{Crack Detection Models with CNN Detector}
\begin{figure}[h!]
\centering
\includegraphics[width=1.\textwidth]{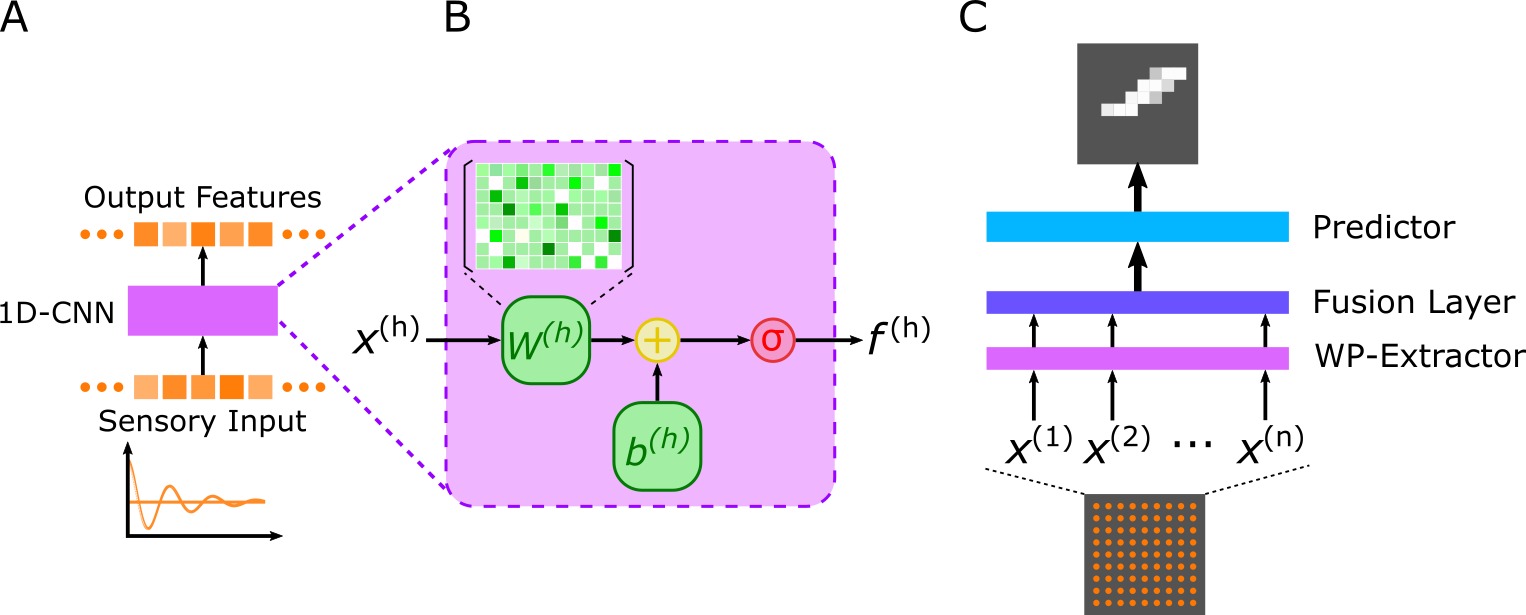}
\caption{Conceptual explanation of the proposed crack detection model with 1D-CNN detector. (A). Diagram of a 1D-CNN that transforms a discretized wave input and produces a discrete feature sequence. (B). Internal structure of a 1D-CNN layer, consisting of a trainable weight $W^{(h)}$, a bias $b^{(h)}$. Activation function is represented by $\sigma$.(C) Diagram of the complete structure of the proposed model.}
\label{fig:schematic}
\end{figure}

The proposed damage detection model is trained to detect the exciting crack in the 2D plate and the damage location on hand of the wave field pattern. The training covers a series of wave fields in randomly generated plates with or without cracks. The receivers are placed inside the plate and along the free boundaries to record the coordinate-depending wave field time histories. Here, at the receiver points, the displacements in x- and y-directions are recorded. These typical 1D Euclidean time series are the data base for the 1D convolution filters within the feature extractors. The proposed deep network model consists of three components: a set of 1D-CNN layers acting as a wave pattern (WP) extractor to select WPs from input displacement time histories; two fully convolutional layers to handle the time dimension and on the receiver's dimension to fuse wave patterns; and a predictor module taking the fused features and making predictions of crack existence (Fig \ref{fig:schematic}).

Let $N$ denote the number of receivers that are placed on the plate, they record displacements for $T$ time steps after a load excitation is added. The reading of the $i^{\mathrm{th}}$ receiver at time $t$ is denoted as $s_k^{(t)}$, where $k \in {1, 2, \cdots, N}$ and $t \in {1, 2, \cdots, T}$. The surface of the plate is decomposed into regular grids. Each cell within the grid covers a small area of the plate surface. The number of cells indicates the spatial resolution of the prediction model. The size of the cell was chosen as 10 times smaller than the wave lengths. Finer resolution requires a larger number of cells and each cell covers a smaller area on the sample surface. In contrast, a coarser resolution results in less number of cells and larger coverage for a single cell. The resulting cells can be denoted by the column and row index as $c_{i, j}$. The model makes binary classification on every grid cell to decide whether damage exists in the cell. To summarize, the model can be written as the following equation, where $f$ presents the proposed model and $\theta$ are all trainable parameters; $p_{i, j}$ is the probability of damage existence in $c_{i, j}$.
\begin{equation}
    f_\theta(s_1^{(1)}, \cdots, s_k^{(t)}, \cdots,  s_N^{(T)}) = \{p_{i, j}\}
\end{equation}

\paragraph{CNN} is a feed-forward network that consists of trainable multistage feature extractors. The feature extractors are trained in an optimization procedure, where the gradient of an objective function with respect to the weights of the feature extractors is calculated and back-propagated \citep{rumelhart1986learning}.

CNN is particularly useful to analyse natural signals which can be represented in arrays of different dimension modalities, sequential signals including language and audio as 1D arrays; images as 2D; and video and volumetric images as 3D. The core operation of CNN is to calculate the convolution of input signals with a set of trainable filters. The transformation produces different kinds of features from input signals in spatial (e.g., images) or temporal (e.g., audio) domain \citep{lecun1995conv}.

In the implementation, CNN differentiates itself from other ANNs by using the local connection and weight sharing strategy. In CNNs, one ``neuron'' connects locally with only a restricted number of ``neurons'' in its previous layers, and the connection weights are shared among all the ``neurons''. Local connectivity, weight sharing, and pooling, are CNN's three key properties for dealing with natural signal \citep{lecun2015deep}. A typical CNN architecture is composed of firstly some convolutional layers (ConvLayer), and then some more ConvLayer, or fully connected layers. ConvLayer detects local features from the previous layer by transforming the input signal with a set of filters. It produces different kinds of feature maps with respect to its filters, then an activation operation is applied to the feature maps. The non-linear activation functions ``squeeze'' the values of a feature maps into a certain range, mostly $[0, 1]$, $[-0.5, 0.5]$ or $[0, +\infty)$. Sometimes, a pooling layer is used for down-sampling the feature maps by taking local average or local maximum values. The pooling layer merges semantically similar features into a higher level \citep{lecun2015deep}. The Pooling layer can be intentionally replaced by setting a larger stride in the convolution layer \citep{springenberg2015striving}.

\begin{figure}[h!]
\centering
\includegraphics[width=\textwidth]{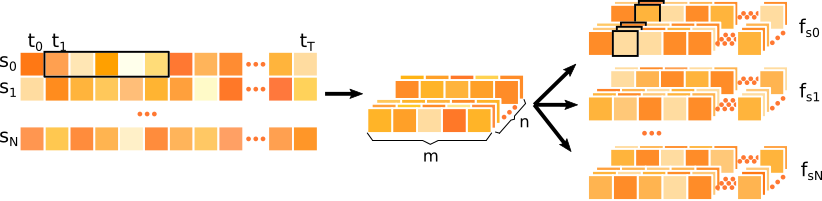}
\caption{The schematic drawing of 1D convolution on N receiver data in T steps.}
\label{fig:conv1d_01}
\end{figure}

Figure \ref{fig:conv1d_01} depicts a schematic drawing of applying \textit{n} 1d-convolution to \textit{N} sensory input of \textit{T} steps. The kernel size is denoted as \textit{m}. For each receiver's input, the convolution produces \textit{n} features. The red mark indicates the data patch used for convolution and the corresponding results in feature maps.
 
\paragraph{The Network Structure} 

The detailed implementation of the proposed model is described in this part. The WP-Extractor consists of three 1D-ConvLayer blocks. Each block has two 1D-ConvLayers followed by a BatchNormalization layer (\citep{ioffe2015batch}) and an activation layer - using the LeakyReLU function. (\citep{he2015delving}). The BatchNormalization layer standardizes the output of the ConvLayer by re-scale according to the samples in a batch. It prevents dramatic change of the spread and distribution of inputs during training, and has the effect of stabilizing and speeding-up the training process. LeakyReLU is a variation of Rectified Linear Unit activation function (ReLU). The ReLU function results in zero when the input is less than zero and keeps the input unchanged when the input is above or equal to zero. The LeakyReLU function ``squeezes'' the value when the input is less than zero and thus allows a small, non-zero gradient when the unit is not active.

The configurations of the two ConvLayers in the same block are identical while the kernel size and number of filters for ConvLayers vary between blocks. To reduce the size of the extracted wave patterns, the output of each 1D-ConvBlock is passed through a MaxPooling layer. Because the 1D-ConvLayers operate only on the time dimension, the WP-Extractor selects only the receiver patterns. For each case, the input data has shape in $N\times T$, where $N$ represents the number of receivers and $T$ the time steps. The output of the WP-Extractor has shape $N\times \hat{T} \times C$, where $C$ is the filter number of the last ConvBlock.

The first fusion layer transforms the receiver data into a 1D vector by passing through a fully convolutional layer. The convolution kernel of this layer is $1\times \hat{T}$, so the transformed field of the ConvLayer covers the whole time domain. The resulted data transformations are given by an $N\times 1 \times C^\prime$ array. After the transformed data are reshaped according to the position of the receivers and has the size of an $9\times 9\times C^\prime$ array ($N=9 \times 9$ as shown in Figure \ref{fig:sample_config}). The second fusion layer, a 2D fully ConvLayer is used to save the information of all receivers. The 2D ConvLayer employs a kernel of $9\times 9$ and thus produces an $1\times 1 \times C^{\prime\prime}$ array.

The core module of the predictor is composed of two Transpose-Convolutional layers (TransConvLayer, sometimes called as deconvolution). The two TransConvLayers are used to up-sample the saved transformed data. TransConvLayers are widely used in image generation tasks, such as DCGAN (\citep{RadfordMC15}). The saved transformed data are first reshaped to $4\times 4\times c$, and then passed through the two TransConvLayers. The ready-for-predict data shape is $16\times 16\times 4$. Finally, it's passed through a 2D ConvLayer of $1\times 1$ kernels, and a single output channel. The sigmoid function is used in this ConvLayer as activation to make sure the prediction ranges from 0 to 1. 

\begin{figure}[h!]
\centering
\includegraphics[width=\textwidth]{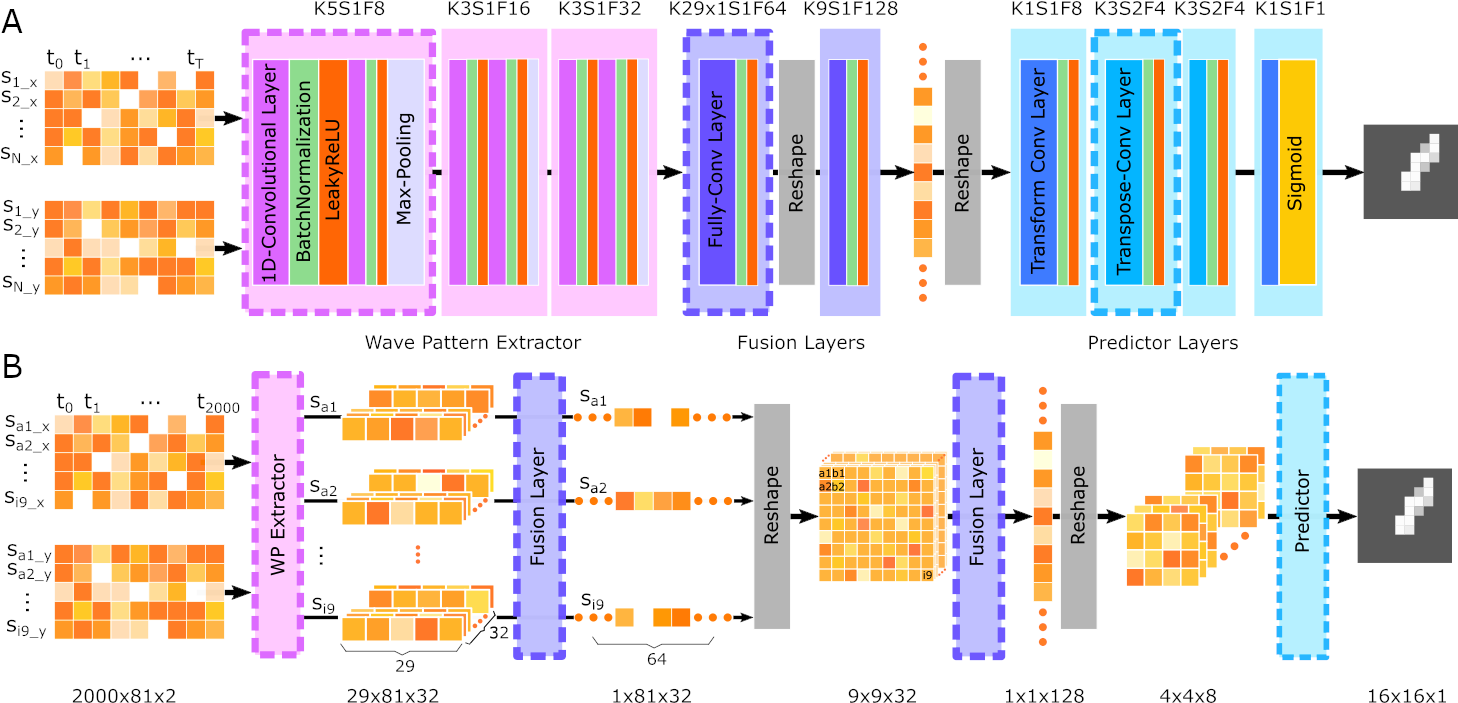}
\caption{Detailed design of the damage detection model. (A). The network architecture. (B) Shapes of input and output of each component.}
\label{fig:model_design}
\end{figure}

The layer configuration and wave pattern shapes are also shown in Figure \ref{fig:model_design}, K refers to the kernel size, S is the steps, F indicates the number of filters. The step of Max-Pooling is set to 4. The 1D convolution can be implemented using 2D convolution by fixing the kernel size of the receiver dimension to 1.

\paragraph{The Loss Function}
Training the proposed model is an optimization procedure, and relies on the objective function (also known as loss function). In this work, the proposed model makes a prediction on crack existence for each single patch. In other words, the model makes multiple binary predictions. For single binary classification problems, cross entropy (CE) loss (see Equation \ref{eq_ce}) is probably the most commonly used loss function. However, considering the ``has crack'' patches actually consist of only a small portion of total patches, cross entropy (CE) loss can introduce bias towards ``no crack'' predictions, i.e., simply predicting all patches as ``no crack'' already result a rather low loss value. To tackle such extreme class imbalance, we select Focal Loss (FL) as our loss function. FL was originally proposed to address the extreme class imbalance in object detection \citep{Lin2017ICCV}. FL is a variation of CE loss by adding a penalty term to reduce the loss value of already correctly predicted training cases. The penalty term $(1-p_t)^{\gamma}$ ($\gamma \geq 0$) re-weights between difficult and easy examples. During training, if a sample is already predicted correctly with a high probability, it is called an ``easy'' case. The penalty term reduces its loss and thus focus on ``hard'' cases, where correct predictions are made with a much lower probability.
\begin{equation}
    L_\mathrm{CE} = -\frac{1}{N}\sum^{N}_{i=1}log(p_t^{(i)}), \quad \mathrm{where} \; p_t = 
    \begin{cases}
     p   &\; \mathrm{if}\;y = 1,\\
     1-p &\; \mathrm{otherwise}
    \end{cases}
    \label{eq_ce}
\end{equation}
where $y$ is the label (1 and 0). 

To adjust the loss values of the two binary classes, a weighting factor $\alpha \in [0, 1]$ can be added. Similar to defining $p_t$, $\alpha_t$ can be defined as $\alpha$ for class 1, and $1-\alpha$ for class 0. The focal loss is written as: 
\begin{equation}
    \mathrm{FL}(p_t) = -\alpha_t(1-p_t)^{\gamma}log(p_t),
    \label{eq_focal_loss}
\end{equation}

For the crack detection case, let $\boldsymbol{y}$ denote the binary image and $\hat{\boldsymbol{y}} = {p_{i, j}}$ a predicted image. The average FL on N cases is calculated by:

\begin{equation}
    \mathrm{FL}(p_t) =\frac{1}{N}\sum_{n=1}^{N}\sum_{i=1}^{U}\sum_{j=1}^{V} -\alpha_t(1-p_{t\{n,i,j\}})^{\gamma}log(p_{t\{n,i,j\}}),
    \label{eq_average_focal_loss}
\end{equation}
where $U$, $V$ are the number of columns and rows of a grid.

Two hyper-parameters are introduced by Focal loss, $\alpha$ and $\gamma$. When $\gamma=0$, FL is equivalent to weighted CE. When $\gamma$ increases, the modulating factor uses a lower standard to define easy examples and has greater power to down-weights well-classified examples. In practice, the optimal $\alpha$ and $\gamma$ are found out by empirical studies.

\subsubsection{Training Process}
\paragraph{Data Pre-processing}

In this simulation, the recorded data at the edge receivers are discarded to avoid any possible effects caused by the extremely large values. Thus, in total the records at 81 receivers are used for both training and testing. Then the wave displacements are normalized between -1 and 1 according to each sample's maximum and minimum value. The resulting input data for the CNN model is $2000\times 81 \times 2$ matrix for each case.

\paragraph{Training Configurations}

The \textbf{Adam} optimization algorithm \citep{KingmaB14} is chosen as the optimizer; it's a commonly used first-order gradient-based optimization algorithm for training deep networks. When updating model parameters during each training step, the algorithm adjusts the learning strength according to the previous gradients. An overview of the optimizers in deep learning can be found in \citep{ruder2016overview}. The initial learning rate is set to 0.0002. The training epochs were set to 150 for all experiments to ensure sufficient training steps for the models to converge. The best model with respect to the evaluation metric is saved for evaluation.

As mentioned in the previous section, the focal loss has two hyper-parameters, $\alpha$ and $\gamma$. To determine suitable hyper-parameters, a set of models were trained with different alpha and gamma. The detailed selection of alpha and gamma is listed in section 3.3.1.

The model and simulation code is implemented in Python with Tensorflow 2.1 Keras. The simulations are performed on a workstation of Windows 10 platform with Nvidia GPU.

\paragraph{Evaluation Metrics}
In this section, we introduce the Dice similarity coefficient (DSC) metric (defined as $M_{DSC}=2\cdot \frac{recall \cdot precision}{recall + precision}$) and the IoU-based accuracy for model evaluation. Although the loss values indicate the quality of the prediction on the patch basis, predicting cracks is more focused. We expect the model to find out the ``has crack'' patches cover as much damaged patches as possible, and as few ``no crack'' patches as possible. 

\begin{figure}[h!]
\centering
\includegraphics[scale=0.6]{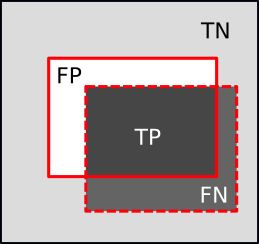}
\caption{Conceptual example of calculating IoU.}
\label{fig:iou}
\end{figure}

In the damage detection study, if the model correctly predicts a cell has (has no) damage, the result is marked as TP (PN); if the model makes wrong predictions on damage existence, the result is either FP or FN as shown in Table \ref{tab:binary_classification}. Based on this, we can calculate precision ($TP/(TP+FP)$) and the recall ($TP/(TP+FN)$), 
\begin{table}
\centering
\begin{tabular}{ c |c |c }
  \hline
       & Right Prediction (T) & Wrong Prediction (F) \\
  \hline
  Has Damage (P)  & TP & FP \\
  \hline
  No Damage (N)  & TN & FN \\
 \hline
\end{tabular}
\caption{\label{tab:binary_classification}Prediction typology in a binary-classification-based damage detection.}
\end{table}
then the DSC metric can be calculated with Equation \ref{eq_dsc}. Similarly, the IoU calculates the ratio between the intersection and the union of ground truth damage area (TDA) and predicted damage area (PDA). As illustrated in Figure \ref{fig:iou}, TDA is bound by dashed red lines;  PDA is marked by solid red lines. Their intersection is the TP set, where the model makes correct predictions. The remaining part of PDA is wrongly identified as damaged area, i.e., the FP set; the remaining part of TDA that has damage inside but has not been found forms the FN set. For a single case, the IoU can be calculated as Equation \ref{eq_iou} by counting the number of each set.

Both DSC and IoU metric range between 0 and 1. If there is no overlap between PDA and TDA, both metrics are equal to 0. When PDA is closer to TDA, the intersection area becomes larger and the union area becomes smaller, resulting in a value closer to 1. When PDA covers TDA exactly, DSC and IoU metrics reach their upper limit of 1. When the sample has no damage and the model makes correct predictions, both the intersection and the union become 0. In this special case, both DSC and IoU metrics are assigned the value 1.
\begin{equation}
    M_\mathrm{DSC}=2\cdot\frac{\mathrm{area}(TP)}{2\cdot\mathrm{area}(TP) + \mathrm{area}(FP) + \mathrm{area}(FN)}
    \label{eq_dsc}
\end{equation}

\begin{equation}
    \mathrm{IoU}=\frac{\mathrm{area}(TP)}{\mathrm{area}(TP) + \mathrm{area}(FP) + \mathrm{area}(FN)}
    \label{eq_iou}
\end{equation}

One underlying assumption of data generation is that each sample has maximum ``one crack'' inside. Based on the assumption, we can define the accuracy using IoU values. For the prediction of a single case, we can consider it as a ``correct'' prediction if its IoU is greater than a given threshold. Given the threshold, the accuracy on the whole dataset is calculated as the ratio of the number of samples whose IoU value is greater than the threshold, to the total number of evaluated dataset.

\section{Results}

\subsection{Simulated Displacement Wave Field Using Dynamic Lattice}

\begin{figure}[h!]
\centering
\includegraphics[width=15cm,height=5cm,keepaspectratio]{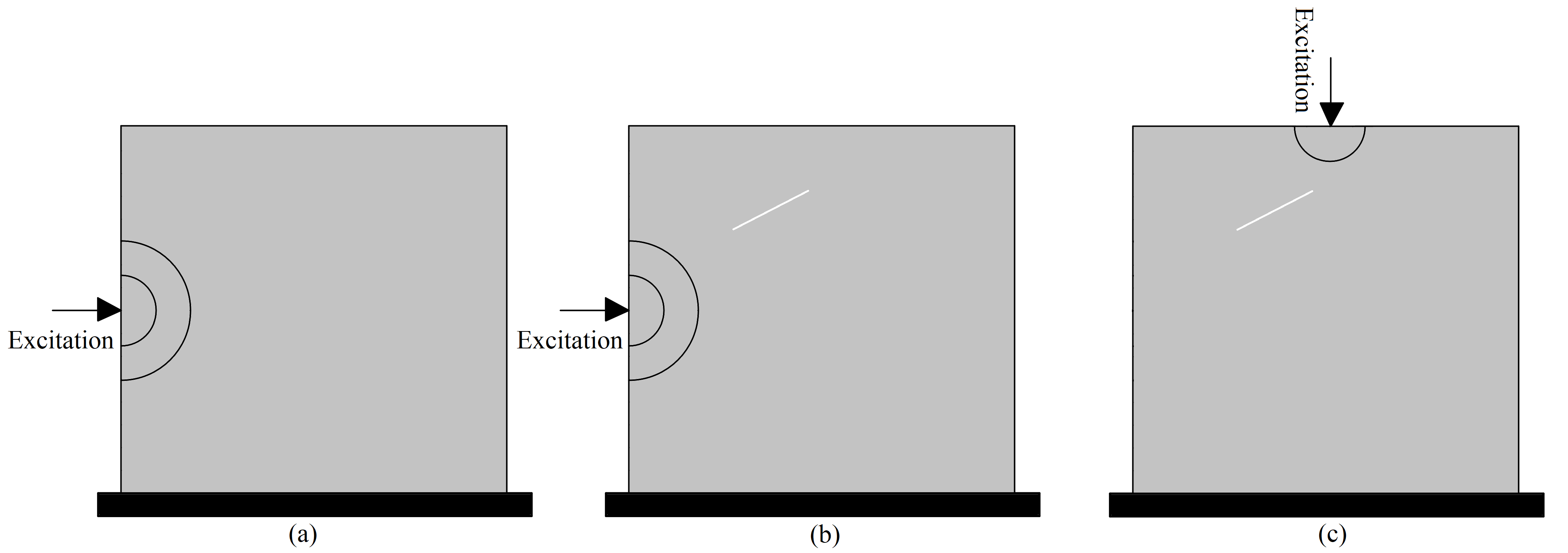}
\caption{Boundary conditions: (a) horizontal excitation without generated crack, (b) horizontal excitation with generated crack, and (c) vertical excitation with generated crack}
\label{fig:Plate_Boundary_1}
\end{figure}

\begin{figure}[h!]
\centering
\includegraphics[width=\textwidth]{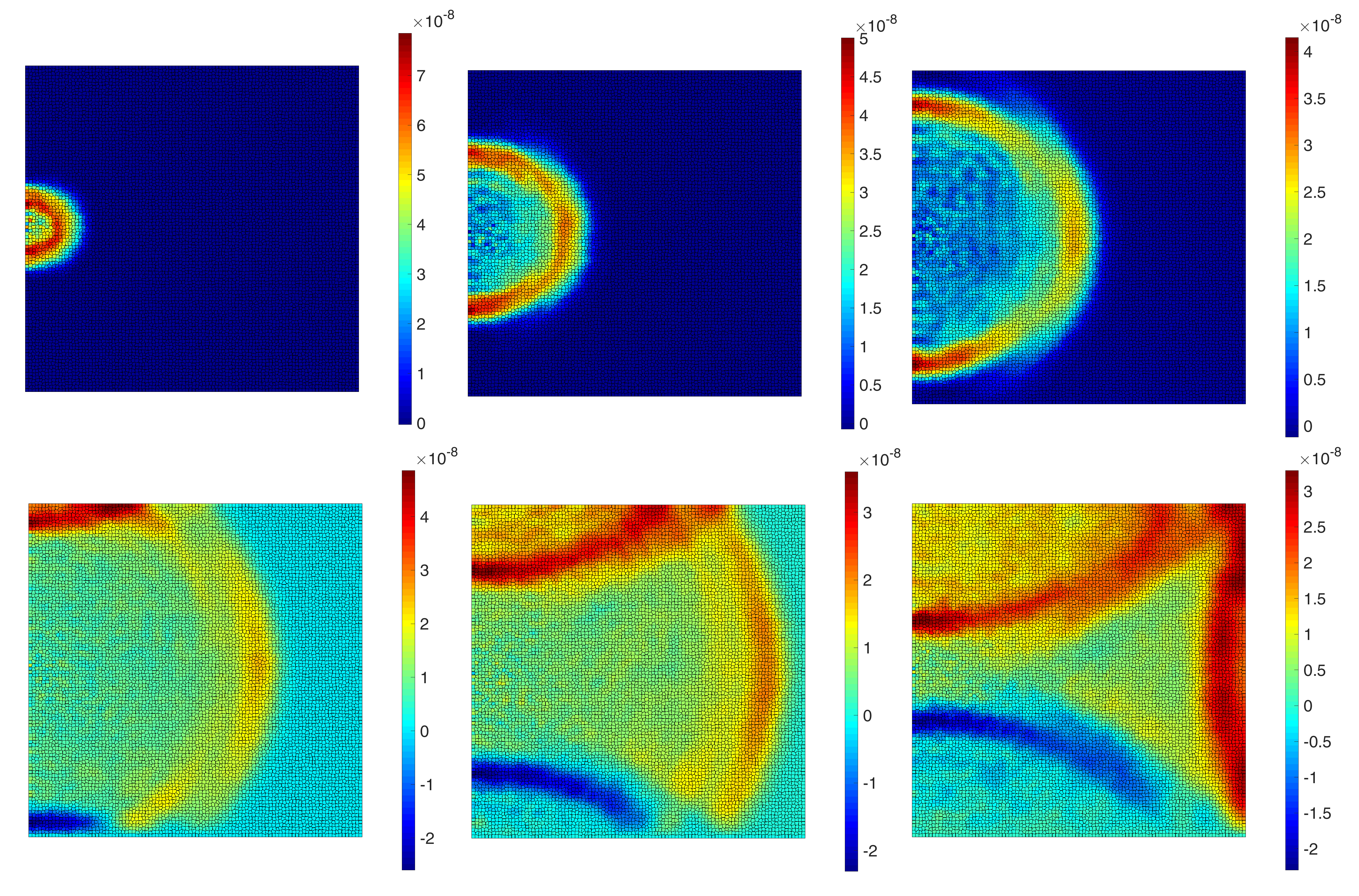}
\caption{The 6 frames (100 time steps interval, from left to right) of a displacement ($u_x$) wave propagation inside the defined plate in  Fig.\ref{fig:Plate_Boundary_1}a.}
\label{fig:nocrack_samp_01}
\end{figure}

\begin{figure}[h!]
\centering
\includegraphics[width=\textwidth]{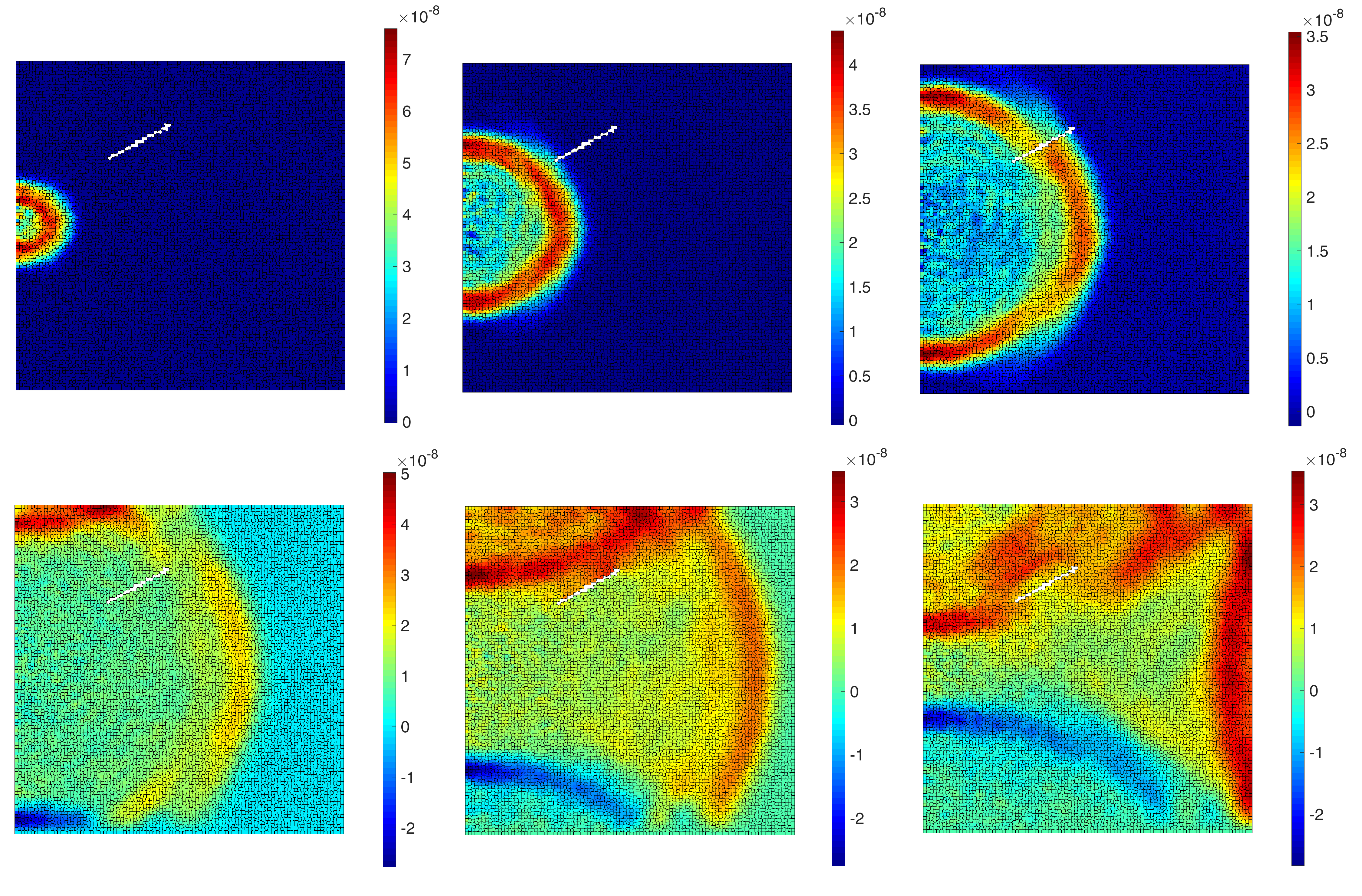}
\caption{The 6 frames (100 time steps interval, from left to right) of a displacement ($u_x$) wave propagation inside the defined plate in  Fig.\ref{fig:Plate_Boundary_1}b.}
\label{fig:crack_samp_02}
\end{figure}

\begin{figure}[h!]
\centering
\includegraphics[width=\textwidth]{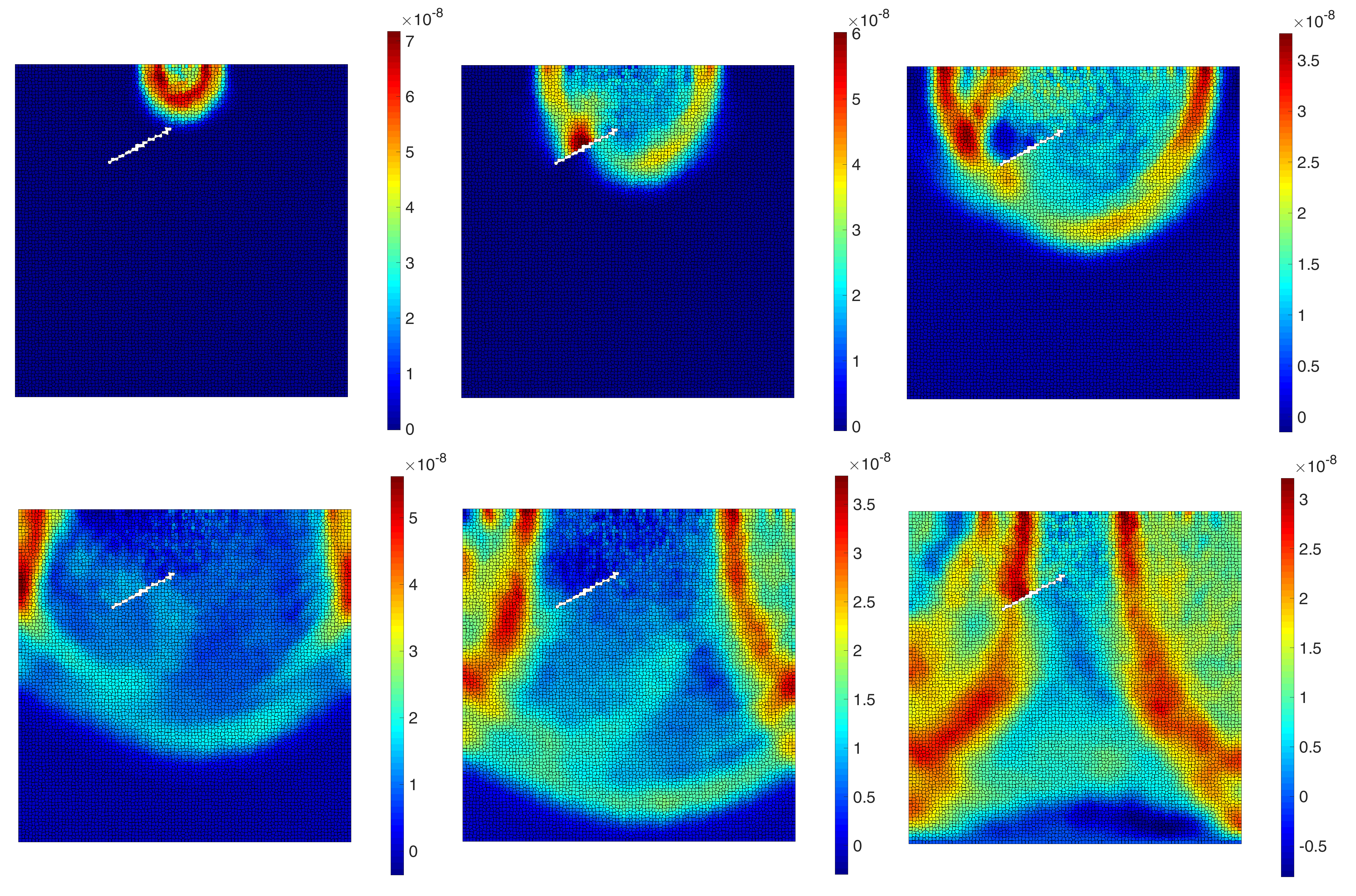}
\caption{The 6 frames (100 time steps interval, from left to right) of a displacement ($u_y$) wave propagation inside the defined plate in  Fig.\ref{fig:Plate_Boundary_1}c.}
\label{fig:crack_samp_03}
\end{figure}

The dynamic lattice method described in section \ref{sec:2_1} is used to simulate the wave fields in a 2D plate. The considered boundary conditions with different excitation points and crack conditions are shown in Fig.\ref{fig:Plate_Boundary_1}. Fig.\ref{fig:nocrack_samp_01} shows the simulated wave fields in lateral direction for a boundary condition of Fig.\ref{fig:Plate_Boundary_1}a. The simulated wave fields with a generated crack (Fig.\ref{fig:Plate_Boundary_1}b) are shown in Fig.\ref{fig:crack_samp_02}. For the conditions shown in Fig.\ref{fig:Plate_Boundary_1}c - excitation point in upper middle boundary - the wave field is plotted Fig.\ref{fig:crack_samp_03}. The results clearly show wave shadows behind the crack as well as the reflection of the wave field from the defined cracked surface. 

\begin{figure}[h!]
\centering
\includegraphics[width=13cm,height=5cm,keepaspectratio]{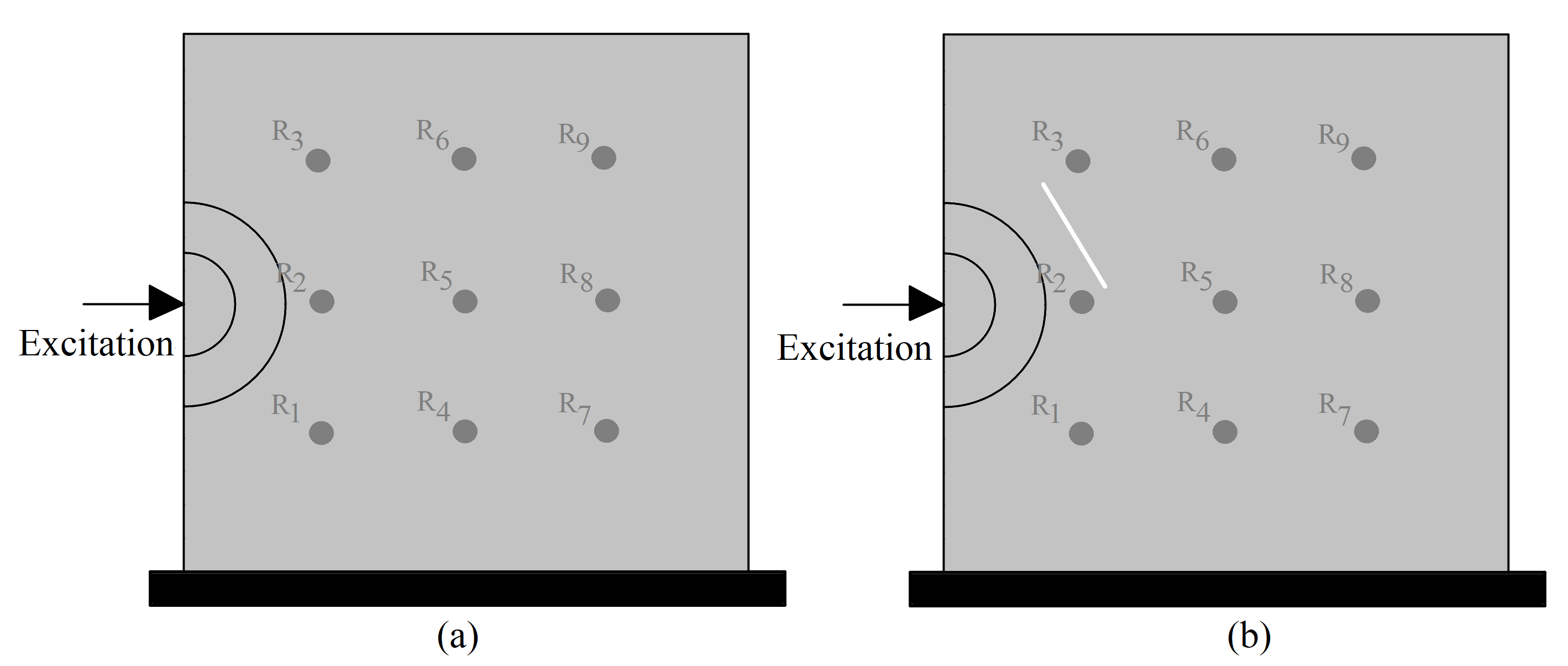}
\caption{The boundary conditions and assigned reference points: (a) without crack, (b) with a generated crack}
\label{fig:Plate_Boundary_2}
\end{figure}

\begin{figure}[h!]
\centering
\includegraphics[width=13cm,height=10cm,keepaspectratio]{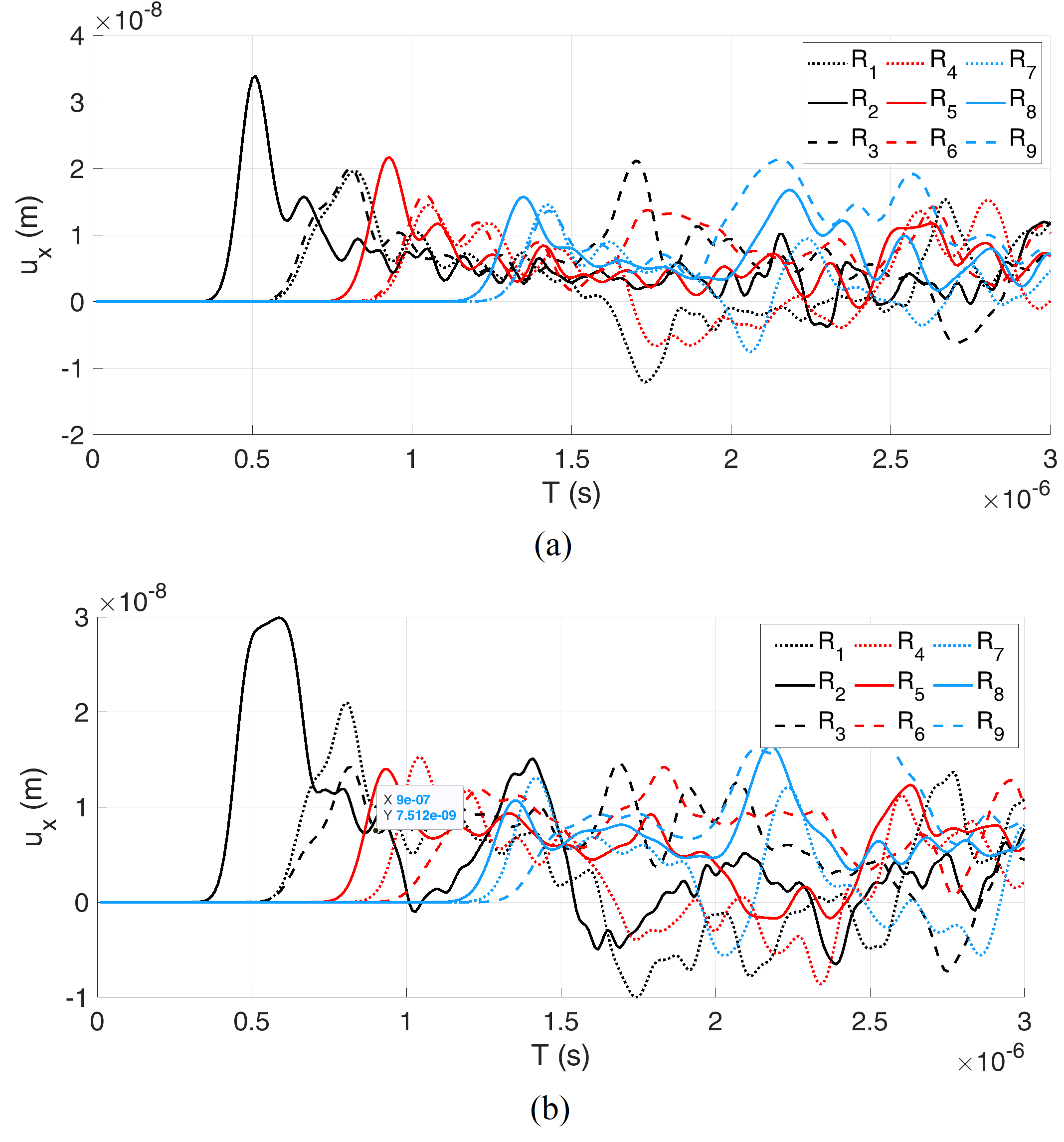}
\caption{The time histories ($u_x$) of the reference points inside the plate: (a) without crack, (b) with a generated crack}
\label{fig:Dis_1}
\end{figure}

The time histories ($u_x$) of the reference points ($R_{1:9}$) inside the plate (Fig.\ref{fig:Plate_Boundary_2}) are shown in Fig.\ref{fig:Dis_1}. Two boundary conditions are considered: one with discontinuity (crack), and one without. The plate dimension is 10x10 $cm$ and the load excitation is at the left middle boundary. The applied rectangular impulse load with a magnitude of 1 $kN$ is kept for 10 time steps, where $\Delta t = 0.00000001 s$. The Young's modulus of a plate is assigned to 5 $GPa$. 

Fig.\ref{fig:Dis_1} clearly shows the arrival time of the wave fields at each reference point. The closest reference point ($R_2$) has the maximum amplitude and minimum arrival time. In Fig.\ref{fig:Dis_1}a the arrival time of the wave field to $R_1\approx R_3$, $R_4\approx R_6$ and $R_7\approx R_9$. Due to the generated discontinuity (crack) in Fig.\ref{fig:Dis_1}b, the first arrival times of the wave field to $R_3$, $R_6$ and $R_9$ are delayed. Theses reference points are located in the shadow field behind the generated discontinuity. Having a closer look at the $R_2$, it is obvious that due to the wave reflection from the generated discontinuity, the arrival of the second wave field happens sooner than the first boundary condition, approximately $1.45\times 10^{-6} s$. The length, location and orientation of the discontinuities affect the wave field in the domain. The simulated wave fields at the reference points are used for training and developing the artificial neural network model.

\subsection{The Trained Damage Detection Model}
\subsubsection{Identified Optimal Hyper-parameters for Focal Loss}
\begin{figure}[h!]
\centering
\includegraphics[scale=0.8]{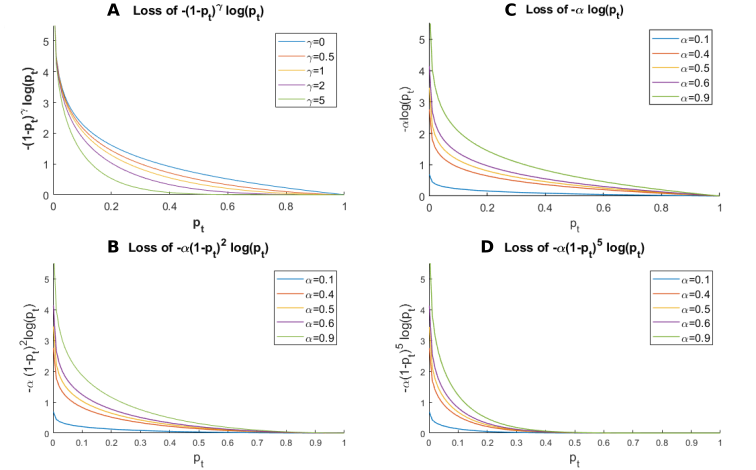}
\caption{The focal loss values with different $\gamma$s and $\alpha$s. (A). Focal loss without weight $\alpha$. (B). Focal loss with different $\alpha$ value with fixed $\gamma=0$. (C). Focal loss with different $\alpha$ value with fixed $\gamma=2$. (D). Focal loss with different $\alpha$ value with fixed $\gamma=5$}
\label{fig:focal_losses}
\end{figure}

FL introduces two hyper-parameters $\gamma$ and $\alpha$. They are used to adjust the loss value of a prediction during the training, so the training can focus on specific types of training cases. Figure \ref{fig:focal_losses} shows the FL value with different $\gamma$ and $\alpha$ values. Figure \ref{fig:focal_losses} (A) is a remake of Figure 1 in \citep{Lin2017ICCV}. With use of the penalty term, the loss value is reduced with the probability of making correct predictions increasing. $\gamma$ controls the decay strength, and larger $\gamma$ ensures the loss to decrease faster. For example, when $\gamma=5$, the predictions where $p_t>0.4$ can hardly contribute to the loss. In contrast, when $\gamma=1$, the predictions where $p_t>0.6$ still contribute to the loss. Meanwhile $\alpha$ can also be used to re-weight the binary classes (has crack and has no crack) (Figure \ref{fig:focal_losses} (B)-(D)). When $\alpha$ is used for one class, the other class is re-weighted by $(1-\alpha)$. Choosing a small $\alpha$ for a class will obviously decrease the contribution of the whole class to the loss. For example, if $\alpha=0.1$ is chosen for the ``has crack'' class and $\gamma=5$, the predictions can be hardly improved when it is greater than $0.3$. Particularly, when $\gamma = 0$ and $\alpha = 0.5$, the FL is equivalent to the cross-entropy (CE) loss.

In this paper, $\gamma$ and $\alpha$ control the model's learning strength on ``no crack'' class and ``has crack'' class. As $\alpha$ increases, the model is driven to focus on damaged cells, because the false predictions of damaged cells contributes more to the overall loss. As $\gamma$ increases, the model is trained to focus on "hard" cells, where the model can't make predictions with high confidence. Because a higher $\gamma$ value forces the model to pay more attention to the "hard" cells. The evaluations of models that are trained with different $\alpha$s are given in Table \ref{tab:results_ce_loss}. When assigning larger weights(larger $\alpha$) to the ``has crack'' class for CE loss, the trained model tends to have lower precision and higher recall. This can be interpreted as the model's tendency to give more ``has crack'' predictions. On the contrary, using small weights(smaller $\alpha$) for the ``has crack'' class results in higher precision but lower recall. This means the models tend to give less ``has crack'' predictions. When $\alpha = 0.9$, the CE loss gives the model of the highest accuracy, however, the model is also characterised as having low precision and high recall; having high DSC metric value but not the optimal one. The results using focal loss are shown in Table \ref{tab:results_fl_loss}. By adding the penalty term, with carefully chosen $\gamma$, the trained models have balanced the precision and recall, and thus result in  an increasing in IoU and DSC metrics. The accuracy is also improved compared to the models trained with CE loss. We consider two sets of $\alpha$ and $\gamma$ combinations, $\alpha=0.35$ $\gamma =0.2$ and $\alpha=0.9$ $\gamma =0.4$, have balanced precision and recall, and achieve high DSC and accuracy at the same time. 

\begin{table}
\centering
\begin{tabular}{ l |l |l |l |l |l }
  \hline
    $\alpha$ & prec. & recall & IoU & DSC & accu.\\
  \hline
  0.1 &	0.852&	0.606&	0.549&	0.709&	0.600\\
  0.2 &	0.917&	0.666&	0.629&	0.772&	0.697\\
  0.25&	0.864&	0.603&	0.551&	0.710&	0.609\\
  0.3 &	0.895&	0.645&	0.600&	0.750&	0.672\\
  0.35&	0.906&	0.705&	\textbf{0.656}&	\textbf{0.793}&	0.706\\
  0.5 &	0.882&	0.675&	0.619&	0.765&	0.666\\
  0.75&	0.756&	0.762&	0.611&	0.759&	0.706\\
  0.9 &	0.839&	0.734&	0.643&	0.783&	\textbf{0.738}\\
  0.95&	0.834&	0.737&	0.643&	0.783&	0.691\\
 \hline
\end{tabular}
\label{tab:results_ce_loss}
\caption{The evaluation results (including IoU, DSC, and accuracy) of models trained by varying $\alpha$ for CE loss($\gamma$= 0).}
\end{table}

\begin{table}
\centering
\begin{tabular}{ l | l |l |l |l |l |l }
  \hline
   $\gamma$ & $\alpha$ & prec. & recall & IoU & DSC & accu.\\
  \hline
0.1& 0.25&	0.916&	0.668&	0.630&	0.773&	0.703\\
0.2& 0.35&	0.880&	0.730&	\textbf{0.664}&	\textbf{0.798}&	\textit{0.756}\\
0.4& 0.9 &	0.803&	0.779&	0.654&	0.791&	\textbf{0.769}\\
1  & 0.5 &	0.884&	0.711&	0.650&	0.788&	0.731\\
2  & 0.75&	0.846&	0.746&	0.657&	0.793&	0.725\\
4  & 0.95&	0.816&	0.748&	0.640&	0.781&	0.722\\
 \hline
\end{tabular}
\label{tab:results_fl_loss}
\caption{The evaluation results (including IoU, DSC, and accuracy) of models trained by varying $\gamma$ for FL loss (optimal $\alpha$).}
\end{table}

\subsubsection{The Selected Thresholds}
The accuracy is calculated dependently with two threshold settings: the threshold of crack existence in a pixel, and the threshold of correct predictions of a sample. The first threshold defines the probability value, above which a pixel can be considered to have a crack inside. In this work, it is also referred to as binarizing threshold ($T_{bin}$). The second threshold is set as a ``tolerance'' (marked as $T_{tol}$) to the prediction. The ``tolerance'' allows a prediction to be ``correct'' when the predicted ``has crack'' pixels cover a certain area of the crack, i.e., its IoU score is greater than the threshold. The very strict criteria requires that the predicted ``has crack'' patches cover the true crack-existing area, i.e., the $\mathrm{IoU}=1$, to be a ``correct prediction''.

\begin{figure}[h]
\centering
\includegraphics[width=0.9\textwidth]{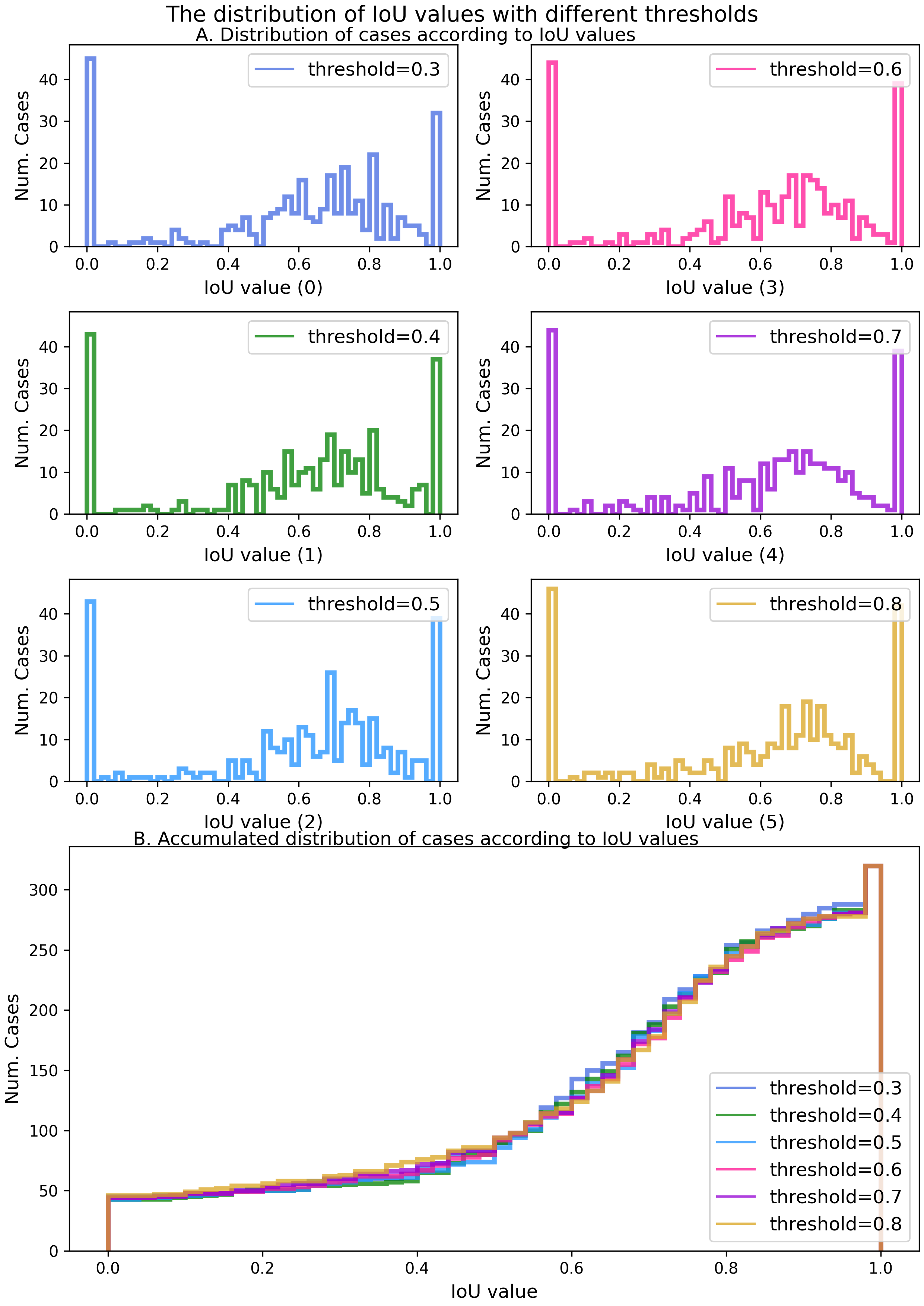}
\caption{The distribution of IoU values among 320 test cases. The IoU values are calculated by the model trained with $\gamma=0.4$ and $\alpha=0.9$. A: the 6 histograms of the IoU values that are calculated with 6 different binarization threshold ($T_{bin}$=0.3, 0.4, 0.5, 0.6, 0.7, 0.8); B: the accumulated histograms of the IoU values that are calculated with 6 different threshold ($T_{bin}$=0.3, 0.4, 0.5, 0.6, 0.7, 0.8).}
\label{fig:histo_iou_a9g4}
\end{figure}

\begin{figure}[h!]
\centering
\includegraphics[width=0.9\textwidth]{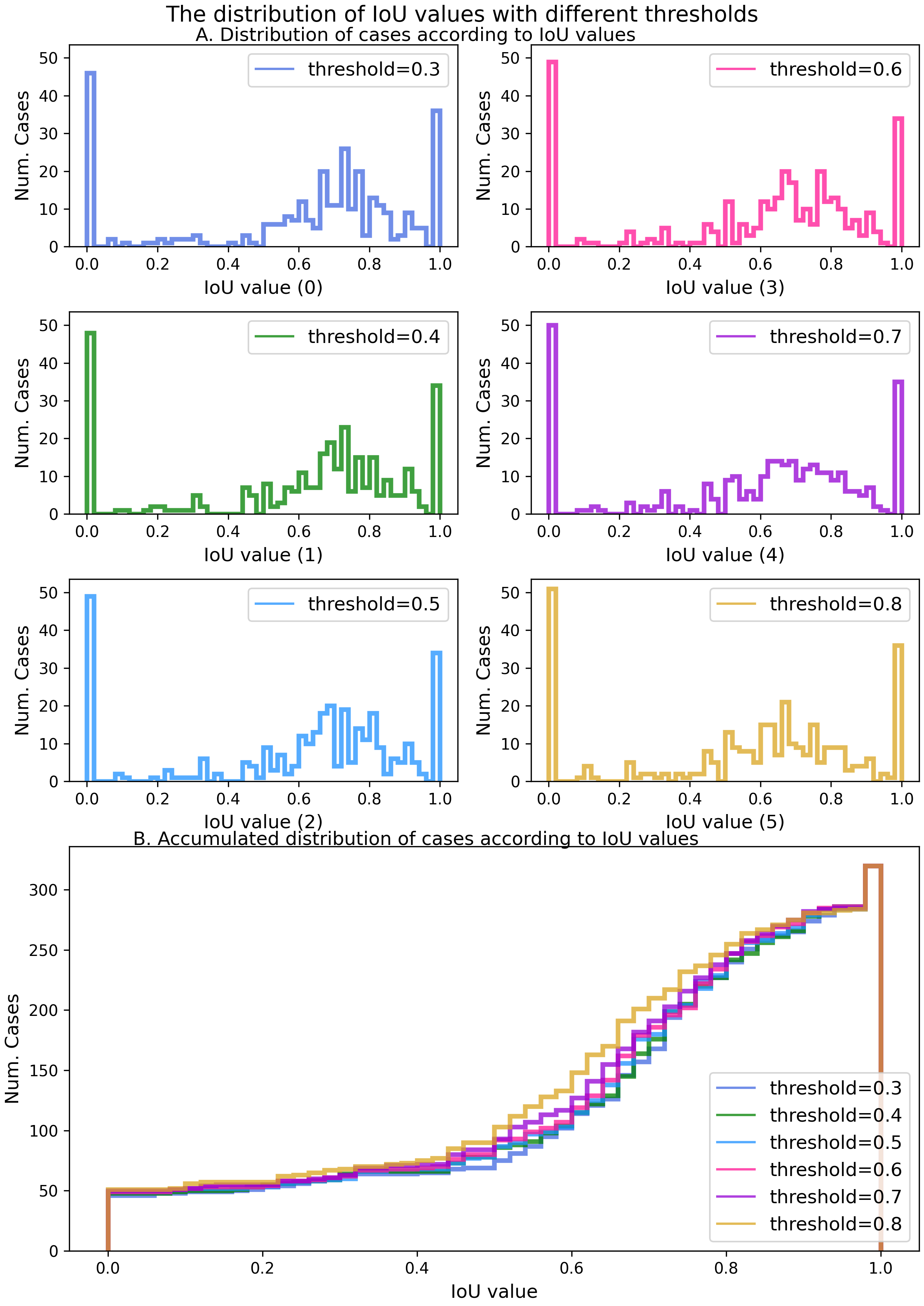}
\caption{The distribution of IoU values among 320 test cases. The IoU values are calculated by the model trained with $\gamma=0.2$ and $\alpha=0.35$. A: the 6 histograms of the IoU values that are calculated with 6 different binarization threshold ($T_{bin}$=0.3, 0.4, 0.5, 0.6, 0.7, 0.8); B: the accumulated histograms of the IoU values that are calculated with 6 different threshold ($T_{bin}$=0.3, 0.4, 0.5, 0.6, 0.7, 0.8).}
\label{fig:histo_iou_a35g2}
\end{figure}

The FL function pushes the predicted probabilities of ``has crack'' and ``no crack'' towards opposite extremes, because a sample with IoU value that is close to 0.5 will have a large penalty during training. This fact is also illustrated in Figure \ref{fig:histo_iou_a9g4} and \ref{fig:histo_iou_a35g2}, which are resulted from the recommended model trained with $\alpha=0.9$ $\gamma =0.4$ and $\alpha=0.35$ $\gamma =0.2$. The sub-figures A of both two figures suggest that most damaged cells are correctly predicted with a probability above 0.5, while still-minor ``hard cases'' get a border prediction around 0.5, with about 45 cases that both models can not properly handle. In both sub-figures B, the accumulated histograms show a clearer comparison on the quality of predictions for different $T_{bin}$ values. They show that different $T_{bin}$ produce similar accumulative histogram curves. This suggests that most "no crack" cells and many "no crack" cells are predicted with very high confidences. We can choose $T_{bin}=0.5$ as it also fits the configuration of FL loss. The curves begin to rise when IoU value reaches 0.5. This suggest us to chose $T_{tol}$ for evaluation, so that the number of cases with IoU values between 0 to 0.5 are relatively small and accumulate quickly when $\mathrm{value}_\mathrm{IoU} > 0.5$. 

\clearpage
\subsubsection{Discussion on Model Performance}
\begin{figure}[h!]
\centering
\includegraphics[width=\textwidth]{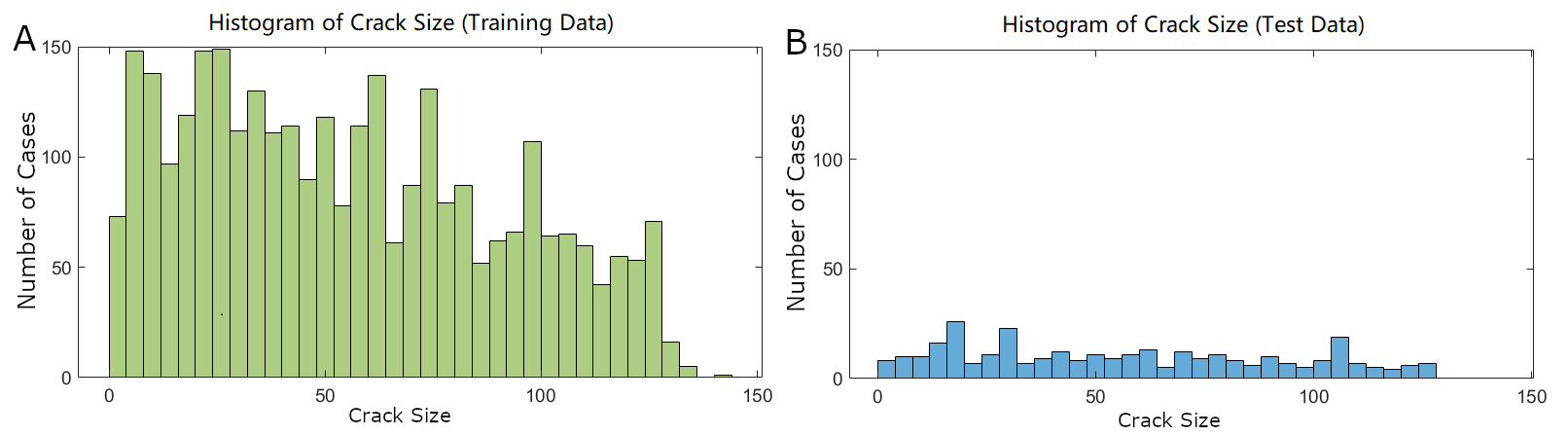}
\caption{Histogram of crack size distribution in training data and testing data. A: crack size distribution in training data; B: crack size distribution in testing data.}
\label{fig:hist_data}
\end{figure}

\begin{figure}[h!]
\centering
\includegraphics[width=\textwidth]{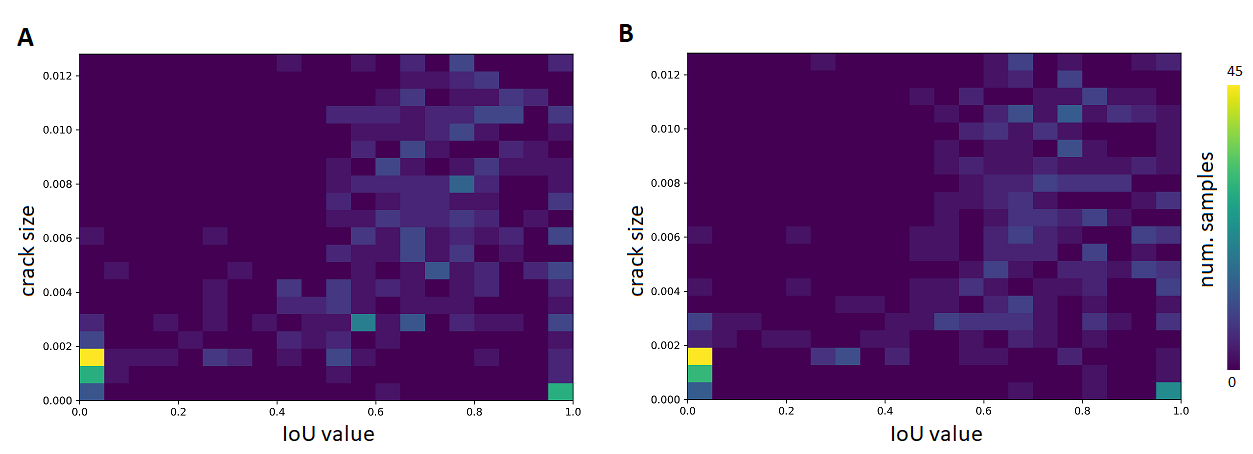}
\caption{Histogram of crack size distribution and IoU values for test data. A: results from model trained with $\alpha=0.9$ $\gamma =0.4$; B: results from model trained with $\alpha=0.35$ $\gamma =0.2$.}
\label{fig:2dhist_crack_iou}
\end{figure}

\begin{figure}[h!]
\centering
\includegraphics[width=\textwidth]{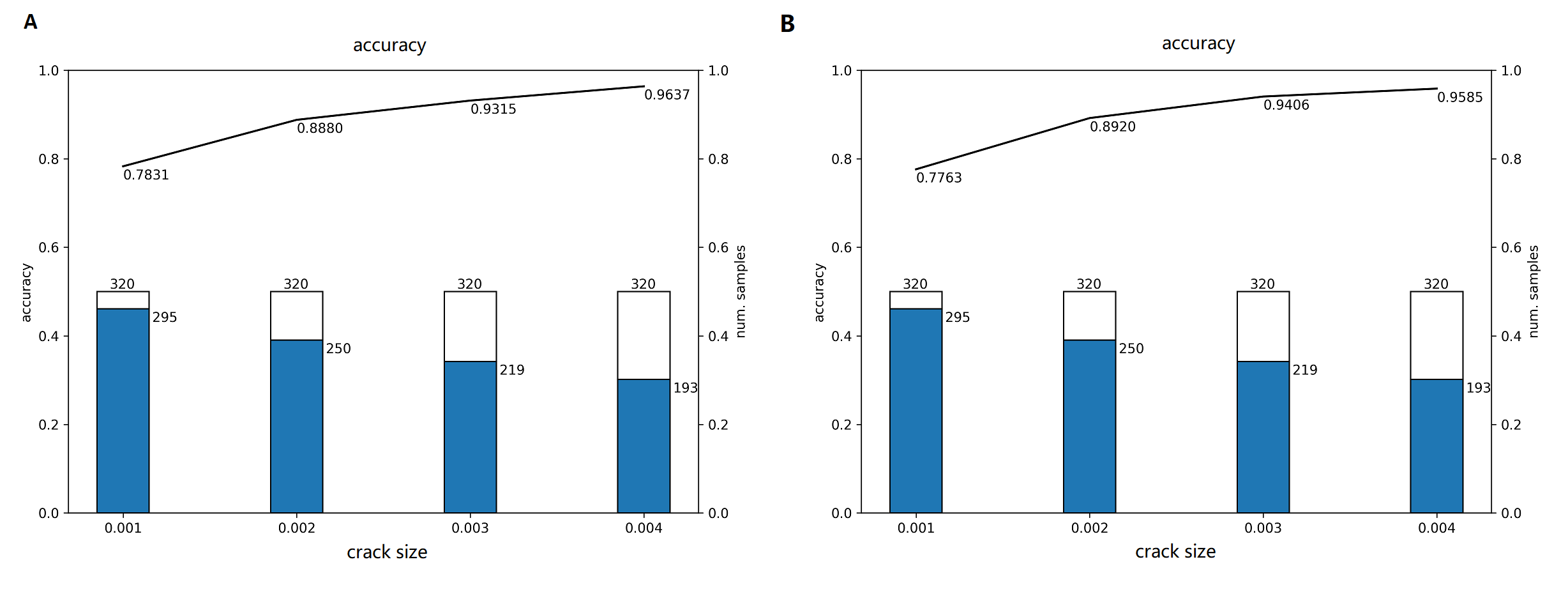}
\caption{The adjusted accuracy calculated for test data after excluding tiny crack cases. A: predictions are made by the model with $\alpha=0.9$ $\gamma =0.4$; B: predictions are made by the model with $\alpha=0.35$ $\gamma =0.2$. The line plot presents the accuracy that is re-calculated when excluding cases with crack size less than 0.001, 0.002, 0.003, and 0.004; the bar chat shows the number of samples after excluding the samples with tiny cracks against the total number of samples in test dataset.}
\label{fig:accuracy_crack}
\end{figure}

The histograms on the distribution of test data (Figure \ref{fig:histo_iou_a9g4} and \ref{fig:histo_iou_a35g2}) indicate that both models are not good at detecting a minor set of damaged cases in test data. We first examine the distribution of crack size (Crack size is the pixel count or the percentage of ``has crack'' pixels in the 100x100 labeling image.) in training data and test data. As shown in Figure \ref{fig:hist_data}, the samples with small cracks consist of a larger portion in test data then in training data. To further explore the relation between crack size and model performance, we plot the histogram of crack size against IoU values for test data (Figure\ref{fig:2dhist_crack_iou}). We can easily find out that the IoU values can be very low for tiny crack samples, whereas the IoU values for larger crack samples are mostly above 0.5. Then the accuracy, adjusted by excluding samples with small crack size from test data, is shown in Figure \ref{fig:accuracy_crack}. It shows that the proposed model is particularly good at identifying larger cracks. IoU values are calculated from the model predictions with $T_{bin}=0.5$. It indicates that the most low-quality predictions are made for samples with tiny cracks, while cases of larger crack sizes generally have better predictions. This means the developed model can easily distinguish between damaged cases and non-damage cases for large cracks but is not good at detecting tiny cracks. If we only count the cases with crack size greater than 0.002, the accuracy leaps by around 0.1. When excluding the cases with small cracks (crack size less than 0.003), the accuracy of the proposed model is already beyond 0.9. If we only count the cases with larger cracks (crack size greater than 0.004), the accuracy of both models can reach 0.95.

\newpage
\section{Conclusion}

The paper presents a new approach to detect damages by wave pattern recognition models. The major development is a learning CNN to detect on-hand the visible wave pattern of the damaged zone within a solid structure. To generate the cracked structure, a new dynamic Lattice Element method was used. The major advantage of this method is the application to heterogeneous structures under mechanical, hydraulically, thermal field influence and local chemical changes to describe the evolution of damages in solid structures. The use of new generation deep CNNs to analyse the time dependency within the changed wave pattern is promising. With the described method, a stable detection of 90 percent of the generated large cracks was possible. The next steps will be the reduction of the used number of receivers and increasing the model's ability of tiny crack detection . 

\section*{Acknowledgement}
The presented work is supported by the Kompetenzzentrum Geo-Energie (KGE) (https://www.kge.uni-kiel.de/, last access: 14 February 2021). We gratefully acknowledge the funding of KGE by the European Union - European Regional Development Fund (EFRE). We also thanks the funding of the project 'Seismic identification of concrete structures' funded by the Federal Ministry of Economic Affairs and Industry - BMWI and the German Federation of Industrial Research Associations - ZIM/ AIF with funding code ZF4016813FF8. Furthermore, we would like to acknowledge the thoughtful reviews of the reviewers and the editor, and their constructive comments supporting the manuscript revision.

\clearpage
\bibliographystyle{plainnat}
\bibliography{references}

\clearpage

\appendix
\section{Example of Predictions}

\begin{figure}[h!]
\centering
\includegraphics[width=\textwidth]{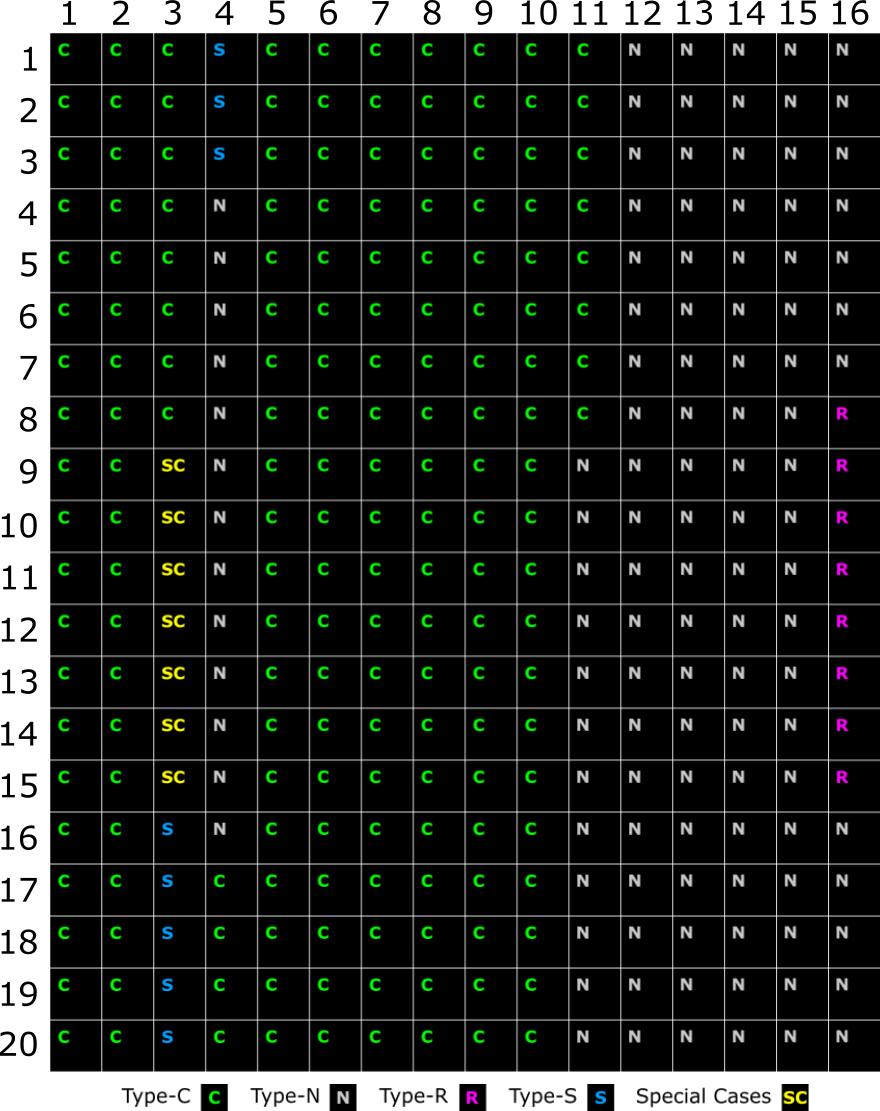}
\caption{The category for 320 test cases. The test cases are categorised into four types: 1). randomly generated samples with randomly generated cracks (\textit{Type-N}), 2). randomly generated samples with no crack (\textit{Type-R}), 3). randomly generated with similar cracks (\textit{Type-S}), and 4). the same sample with different cracks (\textit{Type-C}). They are marked by the colored marks. The special cases (marked as ``SC'' in yellow) are the 7 cases we intentionally generated with the same cracks in training data but from different samples.}
\label{fig:sample-classes}
\end{figure}

\begin{figure}[h!]
\centering
\includegraphics[width=\textwidth]{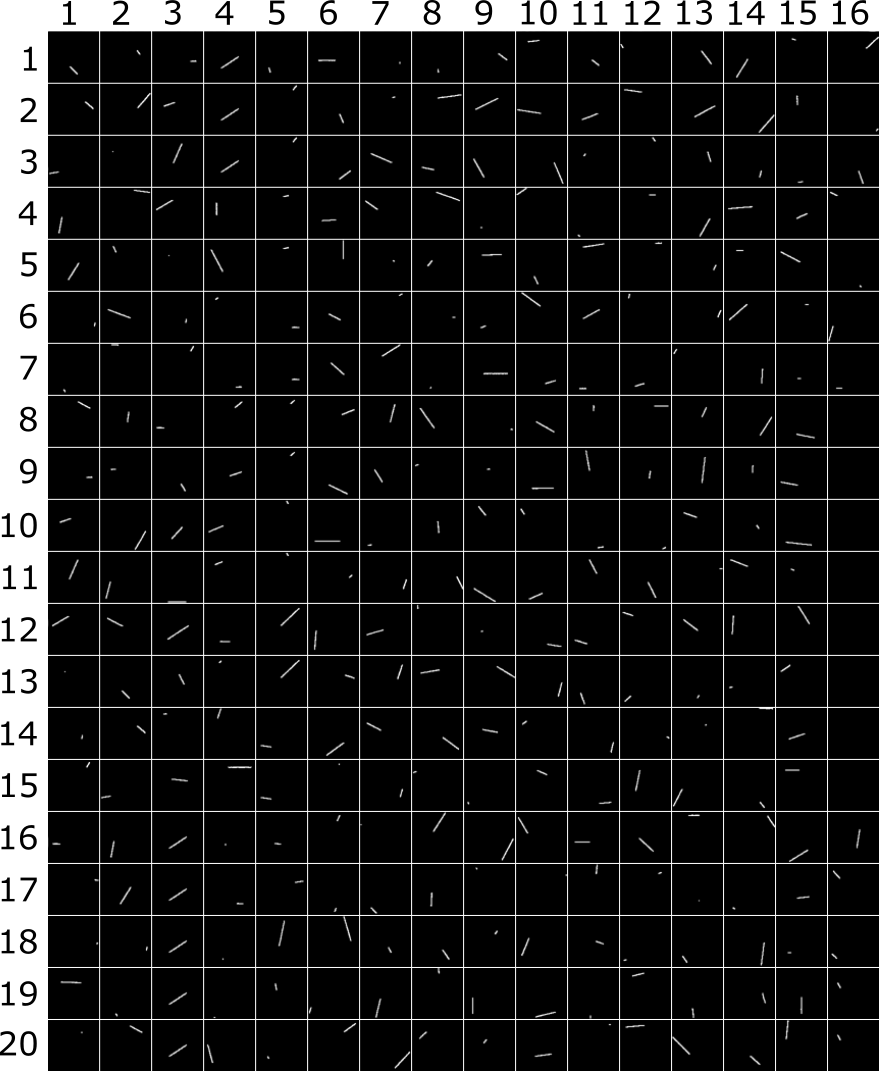}
\caption{The true crack occurrence in reduced resolution (100 x 100) for 320 testing cases.}
\label{fig:groundtruth-valid-origin}
\end{figure}

\begin{figure}[h!]
\centering
\includegraphics[width=\textwidth]{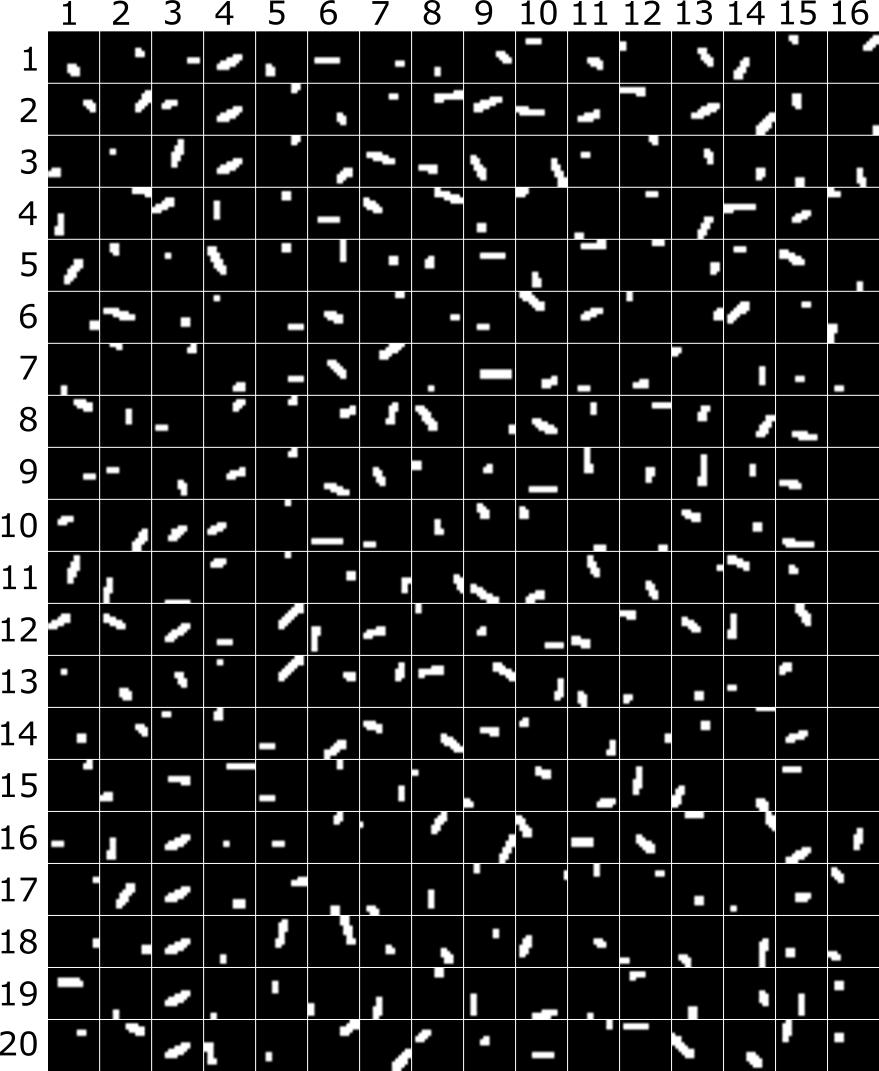}
\caption{The true crack occurrence in reduced resolution (16 x 16) for 320 testing cases.}
\label{fig:groundtruth-valid-reduce}
\end{figure}

\begin{figure}[h!]
\centering
\includegraphics[width=\textwidth]{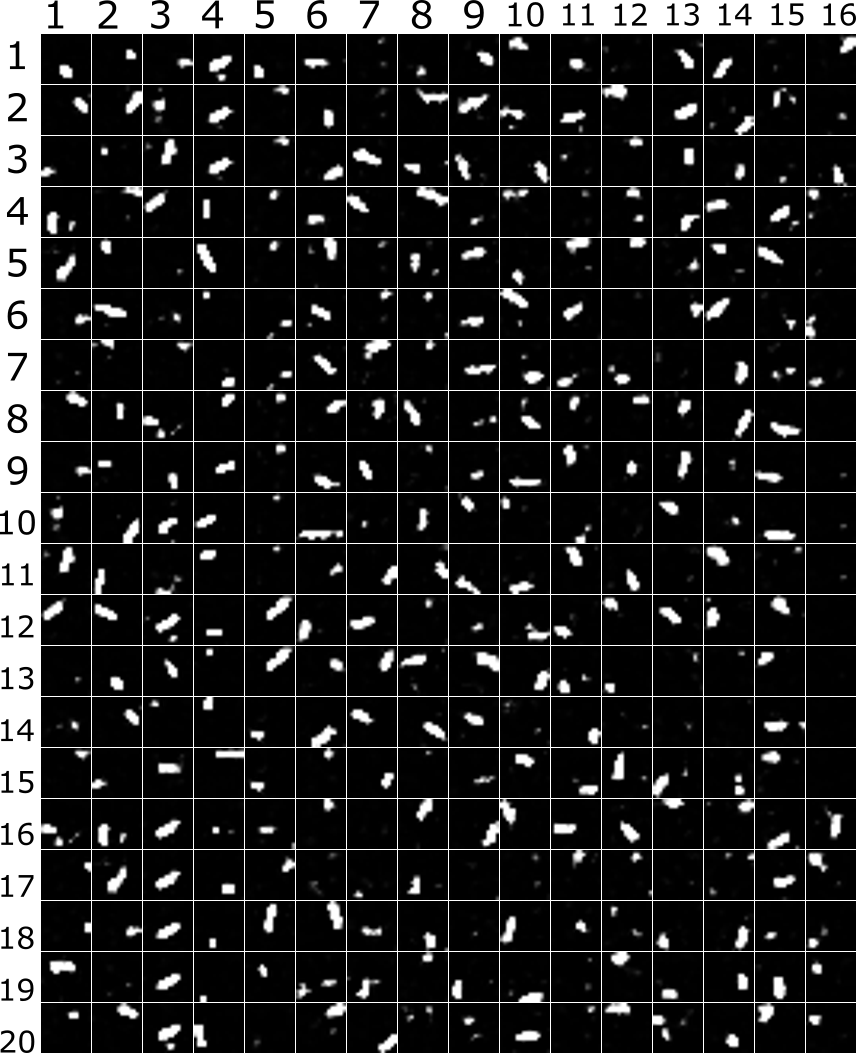}
\caption{The predicted probability of crack existence in pixels for 320 testing cases, the brighter a pixel's color is indicates the higher probability of crack existence inside the pixel. It is made by the model trained with $\alpha=0.9$ and $\gamma=0.4$}
\label{fig:logits-valid-1}
\end{figure}

\begin{figure}[h!]
\centering
\includegraphics[width=\textwidth]{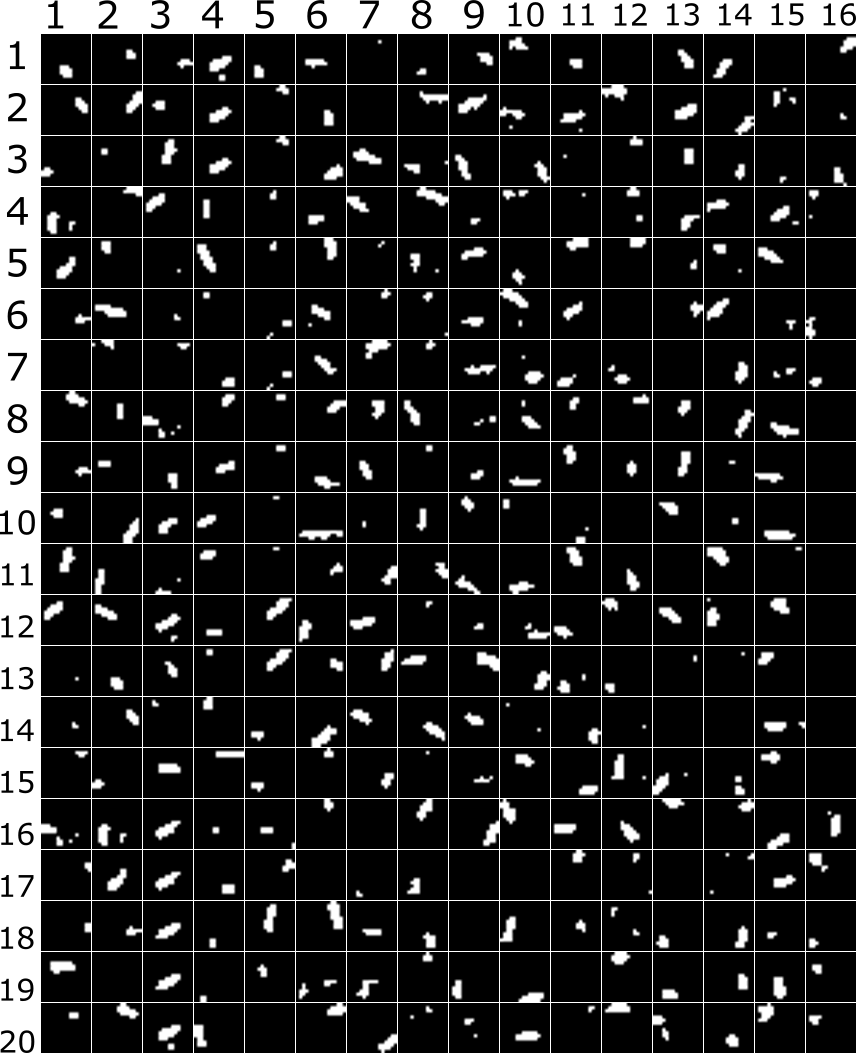}
\caption{The binary predictions made by the model trained with $\alpha=0.9$ and $\gamma=0.4$, where the pixel with a probability greater than a certain threshold (0.5) is considered as having a crack inside.}
\label{fig:pred-valid-1}
\end{figure}

\begin{figure}[h!]
\centering
\includegraphics[width=\textwidth]{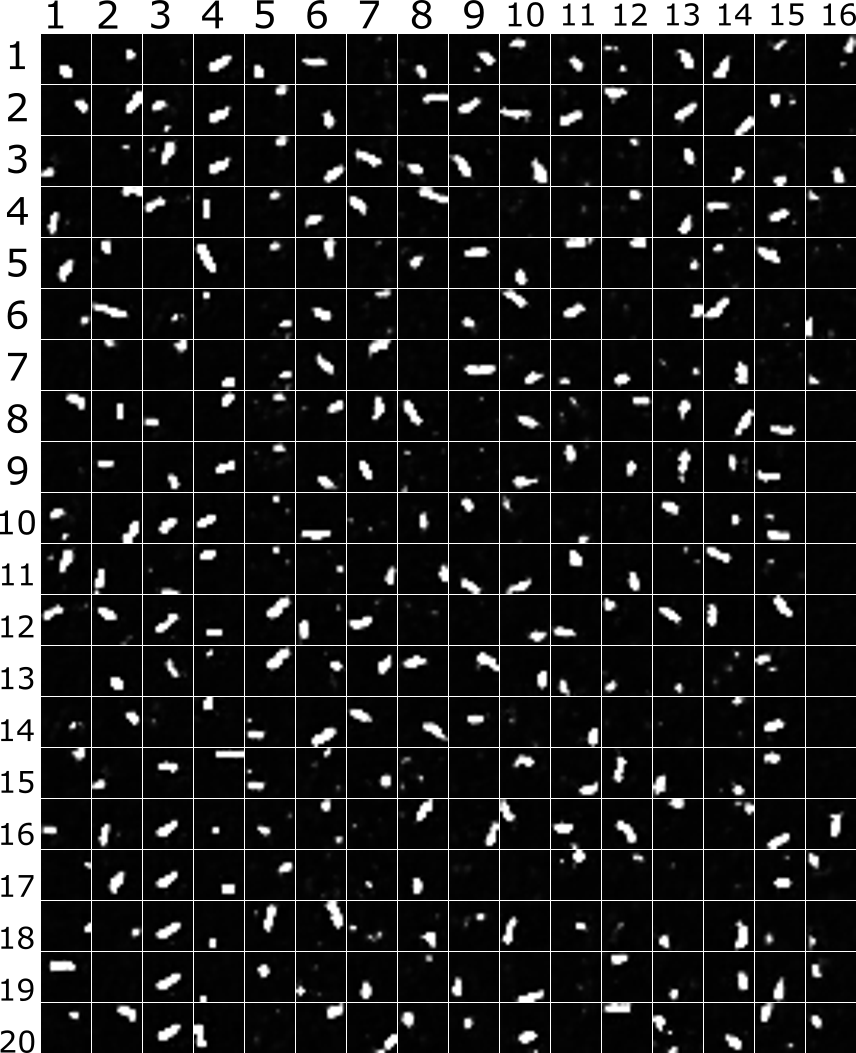}
\caption{The predicted probability of crack existence in pixels for 320 testing cases, the brighter a pixel's color is indicates the higher probability of crack existence inside the pixel. It is made by the model trained with $\alpha=0.35$ and $\gamma=0.2$}
\label{fig:logits-valid-2}
\end{figure}

\begin{figure}[h!]
\centering
\includegraphics[width=\textwidth]{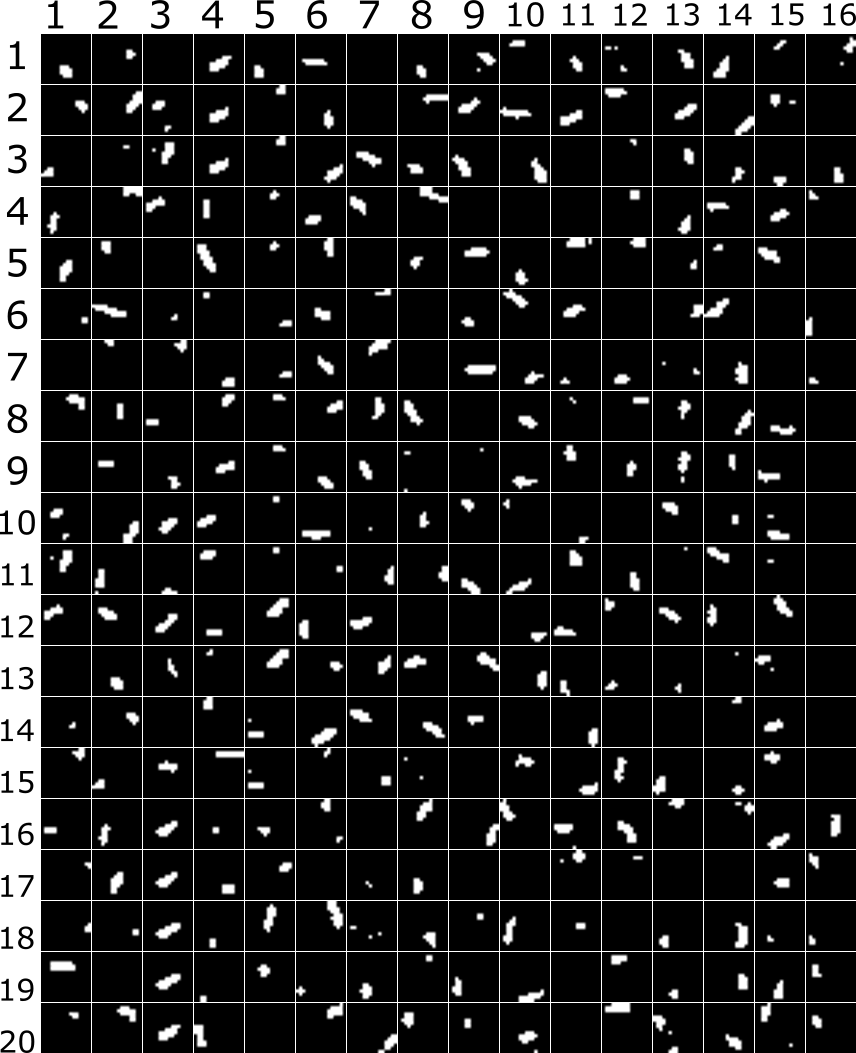}
\caption{The binary predictions made by the model trained with $\alpha=0.35$ and $\gamma=0.2$, where the pixel with probability greater than a certain threshold (0.5) is considered as having a crack inside.}
\label{fig:pred-valid-2}
\end{figure}

The following figures show true crack occurrence and predictions made by our best performing model for 320 valid cases in 16 columns and 20 rows. In Figure \ref{fig:sample-classes}, the 5 different case types (see Section 2.3 ``The Damage Detection Dataset'') are marked with different coloured letters. Figure \ref{fig:groundtruth-valid-origin} is the true occurrence of crack in full resolution (100 x 100) and Figure \ref{fig:groundtruth-valid-reduce} gives the true occurrence in a reduced resolution (16 x 16). The models in this work are trained based on the labels of reduced resolution. The output in Figure \ref{fig:logits-valid-1} - \ref{fig:pred-valid-2} are made by two models which are trained with focal loss with $\alpha=0.9$ $\gamma=0.4$ and $\alpha=0.35$ $\gamma=0.2$, the two recommended alpha and gamma values we found in our experiments. Figure \ref{fig:logits-valid-1} and \ref{fig:logits-valid-2} show the predicted probability of crack existence for each pixel, a brighter pixel indicates a higher probability of crack inside. While in Figure \ref{fig:pred-valid-1} and \ref{fig:pred-valid-2}, binary predictions with the threshold that the pixel with probability greater than 0.5 is considered as having a crack inside.

\end{document}